\documentclass[10pt,twocolumn,letterpaper]{article}

\usepackage[pagenumbers]{iccv} 

\usepackage[utf8]{inputenc} 
\usepackage[T1]{fontenc}    
\usepackage{url}            
\usepackage{booktabs}       
\usepackage{amsmath}
\usepackage{amsfonts}       
\usepackage{nicefrac}       
\usepackage{microtype}      
\usepackage{colortbl}
\PassOptionsToPackage{usenames, dvipsnames}{xcolor}
\usepackage{tabularx}
\usepackage{svg}
\usepackage{tcolorbox}
\usepackage{pifont}
\usepackage[fixed]{fontawesome5}
\usepackage{makecell}
\usepackage{tikz} 

\newcommand{\synInc}{2.6}
\newcommand{\dzInc}{1.4}
\newcommand{\acInc}{3.9}
\newcommand{\synMiou}{70.0}

\newcommand{\cmark}{\ding{51}}%
\newcommand{\xmark}{\ding{55}}%

\newcommand{\src}[1]{{#1}_\text{S}}
\newcommand{\trg}[1]{{#1}_\text{T}}

\newcommand{\promptFeat}{v_\text{CLIP}}
\DeclareMathOperator*{\argmax}{arg\,max}

\definecolor{iccvblue}{rgb}{0.21,0.49,0.74}
\definecolor{customgreen}{HTML}{006633}
\usepackage[pagebackref,breaklinks,colorlinks,allcolors=iccvblue]{hyperref}

\def\paperID{***} 
\def\confName{ICCV}
\def\confYear{2025}
\newcommand{\greenbf}[1]{\bf{\textcolor{customgreen}{#1}}}

\title{LangDA: Building Context-Awareness via Language for Domain Adaptive Semantic Segmentation}

\author{%
    Chang Liu$^{1}$ 
    \quad Bavesh Balaji$^{1}$
    \quad Saad Hossain$^{1}$
    \quad C Thomas$^{2}$ 
    \quad Kwei-Herng Lai$^2$ \\
    \quad Raviteja Vemulapalli$^2$ 
    \quad Alexander Wong$^{1, 2}$ \quad 
     \quad Sirisha Rambhatla$^1$ 
    \\ $^1$University of Waterloo
    \quad $^2$Apple \\
    \texttt{\{chang.liu,bbalaji,s42hossa,sirisha.rambhatla\}@uwaterloo.ca} \\
    \texttt{\{c.thomas,khlai,r\_vemulapalli,alex\_wong3\}@apple.com}
}

\begin{document}

\maketitle
\begin{abstract}

\noindent Unsupervised domain adaptation for semantic segmentation (DASS) aims to transfer knowledge from a label-rich source domain to a target domain with no labels. Two key approaches in DASS are (1) vision-only approaches using masking or multi-resolution crops, and (2) language-based approaches that use generic class-wise prompts informed by target domain (e.g. "a \{snowy\} photo of a \{class\}"). However, the former is susceptible to noisy pseudo-labels that are biased to the source domain. The latter does not fully capture the intricate spatial relationships of objects -- key for dense prediction task. 
To this end, we propose LangDA. LangDA addresses these challenges by, first, learning contextual relationships between objects via VLM-generated scene descriptions (e.g. "a pedestrian is on the sidewalk, and the street is lined with buildings."). Second, LangDA aligns the entire image features with text representation of this context-aware scene caption and learns generalized representations via text. With this, LangDA sets the new state-of-the-art across three DASS benchmarks, outperforming existing methods by 2.6\%, 1.4\% and 3.9\%.

\end{abstract}  
\section{Introduction}
\label{sec:intro}

Semantic segmentation is a dense prediction task that demands expensive and time-consuming pixel-level annotations \cite{cordts2016cityscapes, sakaridis2021acdc}. This problem is further exacerbated by domain shift when additional annotation is required as the data evolves \cite{tranheden2021dacs,hoyer2022daformer,hoyer2023mic}. To alleviate the need for manual annotation, unsupervised domain adaptation for semantic segmentation (DASS) methods train segmentation networks -- usually based on the student-teacher architecture -- on an available labeled source domain and bridge domain gaps by adapting to an unlabeled target domain.
\cite{zou2018unsupervised_class_balanced_self_train,hoyer2022hrda,hoyer2022daformer,vu2019advent,tranheden2021dacs}.

\begin{figure}[t]
    \centering
    {{\includegraphics[width=\columnwidth]{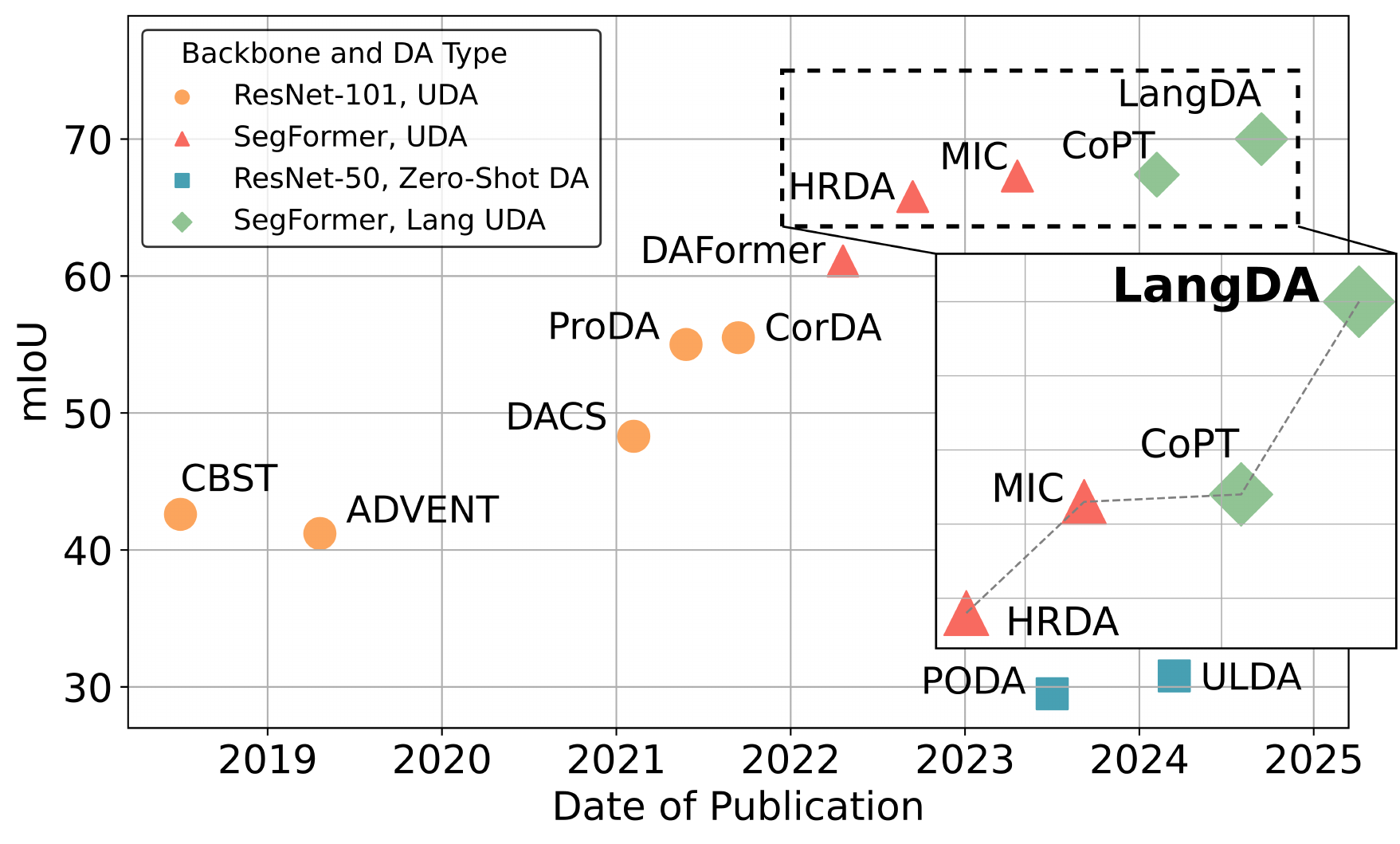}}}%
    \caption{\textbf{Synthia $\to$ Cityscapes: Progress of DASS over time.} 
    Improvements in UDA methods have plateaued in the last two years. Compared to MIC \cite{hoyer2023mic}, which tries to learn spatial relationships only on vision domains, and CoPT \cite{mata2024CoPT}, which employs generic language-priors, our proposed LangDA uses contextual information from descriptive captions, achieving state-of-the-art performance.}
    \vspace{-18px}
\end{figure}

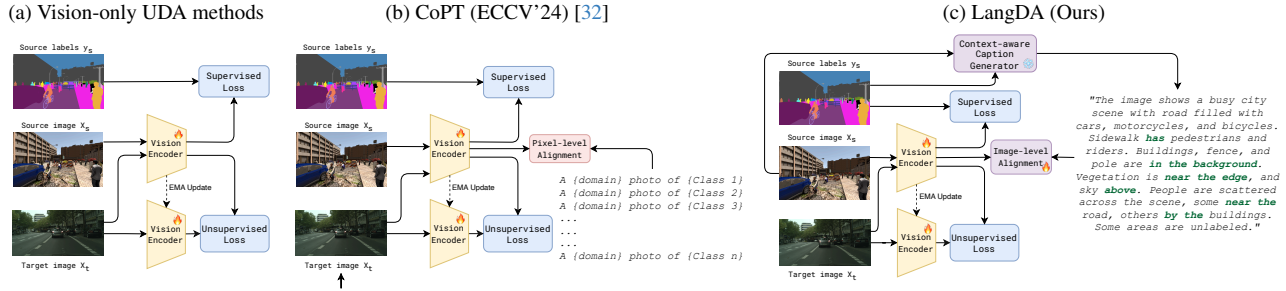
\begin{figure*}[t]
    \centering
    \begin{tikzpicture}
        \node at (0,0) {\includegraphics[width=\linewidth]{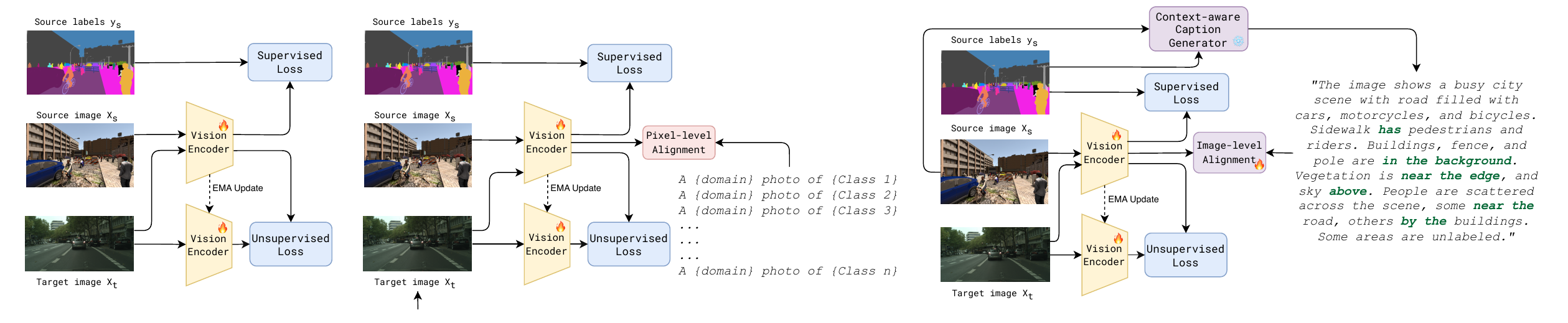}};
        \node[font=\footnotesize] at (-6.8, 2) {(a) \text{Vision-only UDA methods}};
        \node[font=\footnotesize] at (-2, 2) {(b) \text{CoPT (ECCV'24) \cite{mata2024CoPT}}};
        \node[font=\footnotesize] at (5, 2) {(c) \text{LangDA (Ours)}};
    \end{tikzpicture}
  \caption{\textbf{(a)} \textbf{Vision-only UDA} leverages an EMA-updated teacher-student framework with consistency losses to segment unlabeled target data. \textbf{(b)} \textbf{CoPT} uses LLM-generated class-wise text prompts and performs pixel-level alignment (aligns pixel features to corresponding class prompts), not focusing on spatial relationships in language. They also require additional  supervisory text prompts for target domain. \textbf{(c)} Our proposed method \textbf{LangDA} utilizes context-aware image captions and performs image-level alignment (aligns image features to the image captions) to facilitate context-aware domain-invariant adaptation. Words providing context are highlighted in {\greenbf{green}}.}
    \label{fig:model_diagram}
    \vspace{-15px}
\end{figure*}

Capturing \emph{context} -- the spatial relationships between objects, is key to accurate image segmentation \cite{hoyer2023mic}. Existing vision-only unsupervised domain adaptation (UDA) methods \cite{hoyer2022hrda, hoyer2023mic, hoyer2022daformer} (Fig. \ref{fig:model_diagram}a) attempt to build context-awareness by \emph{implicitly} learning spatial relationships between different patches of the image. For instance, to capture fine-grained and long-range context dependencies from an image, HRDA \cite{hoyer2022hrda} uses high-resolution crops along with low-resolution images. On the other hand, MIC \cite{hoyer2023mic} masks target images in the source-trained model.  While these methods provide substantial improvements for DASS, they are susceptible to noisy pseudo-labels biased to the source domain \cite{mata2024CoPT,hoyer2023mic}. To mitigate noisy predictions, CoPT \cite{mata2024CoPT} utilizes generalized representations in the form of language to guide the student network. Specifically, CoPT aligns generic class-level text features with pixel features (e.g. text embedding of "a photo of a \{car\}" with image pixels of a "car"). However, sole alignment of text and pixel features of each class separately (pixel-level alignment) ignores the context relationships between different classes (e.g. "car" and "pedestrian").

Motivated by the above, we propose LangDA, where we explicitly build contextual understanding through text. Unlike existing language-guided methods \cite{mata2024CoPT,huang2023sentence,fahes2023poda,yang2024unified} using generic class-level prompts, LangDA aligns image scenes with descriptive captions (e.g., "a \{pedestrian\} is on the \{sidewalk\}, and the street is lined with \{buildings\}..."; see \Cref{fig:vlm-gen-ex}), effectively providing models with language guidance on object relationships. To fully leverage the context relationships in the caption, we introduce image-level consistency module, bringing the entire image's features closer to the corresponding caption. We demonstrate that leveraging contextual relationships in language allows the model to generalize much better to target data.

Our formulation differs from existing language-guided DASS approaches in three ways. (1) Existing methods rely on supervisory text to describe domain gap (e.g., "\textit{\{daytime\}} photo" to "\textit{\{nighttime\}} photo") \cite{mata2024CoPT}, which fails in real-world scenarios where domain shifts are unpredictable (e.g., tumor scans with color and location variations across hospitals \cite{hu2025tumour_uda,lee2023tumour_self,basak2024tumour_quest}). We address this by using a context-aware caption generator (with seeded VLM and LLM) to \textit{automatically create} image-specific captions, ensuring effectiveness even for unknown domain gaps. (2) We remove the need for manual prompt tuning \cite{huang2023sentence,yang2024unified,fahes2023poda,mata2024CoPT} and human feedback \cite{gong2024coda}, which are often inconsistent, difficult to reproduce, and can be resource-intensive. This standardizes captions for benchmarking, eliminating the effort required for prompt engineering.
(3) Instead of simply matching individual class embeddings (e.g., text embedding of "car" with its pixel embedding), we align a descriptive scene embedding with its corresponding image embedding (image-level alignment), thus encoding richer object relationships via language.

We validate LangDA on three different DASS settings: Synthia $\to$ Cityscapes, Cityscapes $\to$ ACDC, and Cityscapes $\to$ DarkZurich. LangDA achieves state-of-the-art performance in all three UDA settings, surpassing existing methods \cite{hoyer2023mic,mata2024CoPT} by \textbf{\synInc\%}, \textbf{\acInc\%}, and \textbf{\dzInc\%} respectively, for the challenging semantic segmentation tasks with domain shifts. Our ablation studies further highlight the superiority of context-aware image-level alignment over pixel-level alignment. These results confirm LangDA's capacity to extract spatial relationships encoded in language representations for robust domain adaptation. 

\section{Related Works}
\label{sec:related_works}

\begin{figure*}[ht!b]
    \centering
    {{\includegraphics[width=\textwidth]{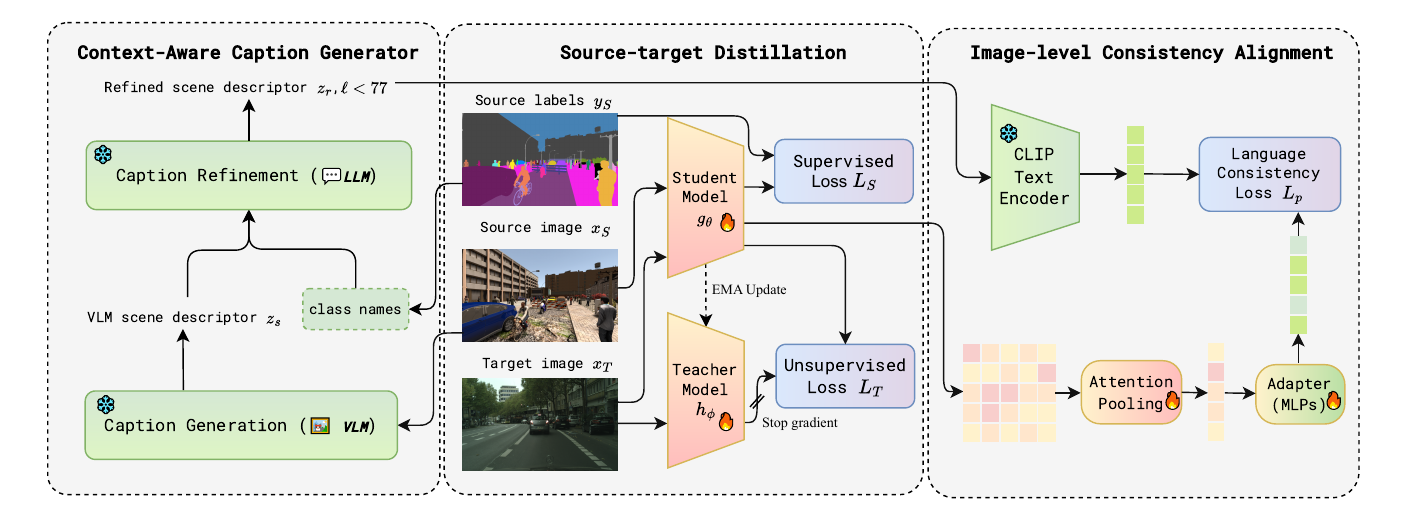}}}%
    
    \vspace{-10px}
    \caption{\textbf{LangDA Architecture}. LangDA is a prompt-driven UDA framework that leverages contextual language descriptions to bridge domain gaps between labeled source images and unlabeled target images. LangDA includes two modules: context-aware caption generation and language-consistency alignment. \emph{Left:} Context-aware generation is a two step process. First, a captioning model generates captions that encode context relationships for the source image (e.g. "there is a sidewalk \textit{on one side of} the street"). Then, the captions are improved by passing class names from ground truth labels into an LLM.
    \emph{Right:} In the image-level consistency alignment module, an adapter (explained in \Cref{method:prompt_adapt}) projects the image features from the trained network onto the same latent space as text embeddings. The LangDA image encoder is trained from scratch because the CLIP image encoder performs poorly on semantic segmentation tasks.} 
    \vspace{-15pt}
    \label{fig:model_arch}
\end{figure*}

\subsection{Unsupervised Domain Adaptation}
In UDA, a model trained on a labeled source domain is adapted to an unlabeled target. Existing UDA methods fall into three categories: (1) discrepancy minimization that reduces domain gap using statistical distance functions \cite{long2017maxmeandiscre, sun2016corr_al,sun2016corr_return, vu2019advent,grandvalet2004semi_entrop_min,wang2021domain,zhang2021prototypical}; (2) adversarial training uses a domain discriminator to promote domain-invariant features \cite{vu2019advent,goodfellow2014gan}; (3) self-training generates pseudo-labels \cite{lee2013pseudo, hoyer2022daformer} and applies consistency regularization \cite{tarvainen2017mean_tea_consis_regu, tranheden2021dacs}. Additional reviews on related works can be found in Appendix \ref{sec:additional_rel_works}.

In particular, self-training has shown strong results in segmentation when attempting to learn spatial relations using self-supervised objective. To better leverage ImageNet's real-world high-level semantic classes, DAFormer \cite{hoyer2022daformer} proposes a transformer-based DASS method and regularizes the bottleneck image features with ImageNet features. HRDA \cite{hoyer2022hrda} uses high-resolution crops in conjunction with low-res crops to improve fine-grained representation, while MIC \cite{hoyer2023mic} leverages masked image modeling to try to implicitly learn spatial relationship in images. Although self-training based methods are more stable and outperform other UDA approaches \cite{hoyer2022daformer}, they remain vulnerable to noisy pseudo-labels dominated by source features, limiting target domain performance \cite{mata2024CoPT}. To further bridge the domain gap, recent methods such as CoPT \cite{mata2024CoPT} turn to CLIP-based models to leverage its generalized world prior. Although CoPT \cite{mata2024CoPT} aligns the covariance matrix of image and text features, they facilitate this alignment at pixel-level and did not take advantage of the spatial relationship between objects in VLM's representation. Our work addresses this research gap by leveraging spatial contextual information in texts to bridge the source and target domain in DASS.

\subsection{Domain Adaptation with Language}

In zero-shot domain adaption for semantic segmentation, recent works adapts to unseen domains using a textual description of the unavailable target data. For instance, PODA \cite{fahes2023poda} utilizes text embedding from the CLIP model to approximate the target visual domain. ULDA \cite{yang2024unified} notes PODA requires separate segmentation heads and adaptation steps for each target domain. Its follow-up work addresses this by enabling a single model to adapt to multiple domain settings.

In UDA, Lai \textit{et al.} \cite{Lai_2023_ICCV} initially formulated a pseudo-labeling setting and active debiasing of CLIP for unsupervised domain-adaptive classification. Following this, Du \textit{et al.} \cite{du2024domain} exploit domain-invariant semantics by mutually aligning visual and text embeddings for domain adaptive classification. Jin \textit{et al.} \cite{jin2024llms} leverage language to compute fine-grained relationships between different parts of an object for UDA-detection. Lai \textit{et al.} \cite{Lai_2024_WACV} design a domain-aware pseudo-labeling scheme for effective domain disentanglement. Lim \textit{et al.} \cite{lim2024cross} uses CLIP for adaptation when source and target domain have different class names. CoDA \cite{gong2024coda} adapts to adverse weather by generating additional images via text-to-image synthesis and requires costly human feedback for image selection. Despite their success, we do not compare existing methods against \cite{gong2024coda}. Since it relies on extra data and manual intervention, comparing it to methods trained solely on existing data would be neither fair nor meaningful. CoPT \cite{mata2024CoPT} aligns the covariance matrix of class-wise text embeddings (averaged from source and target domains) with pixel-level image features for DASS. In contrast, no existing DASS method directly aligns spatial relationships in text with visual representations. Our approach fills this research gap by capturing object relationships in language descriptions, and aligning it with image features to mitigate visual domain discrepancies.

\section{Methodology}
\label{sec:method}

\setlength{\abovedisplayskip}{0.5pt}
\setlength{\belowdisplayskip}{0.5pt}

\subsection{Problem Formulation}

In DASS task, we learn a model \( f_\theta: \mathbb{R}^{H \times W \times 3} \rightarrow [0, 1]^{H \times W \times K} \) parameterized by $\theta$ that performs pixel-wise classification of $K$ classes on an unlabeled target domain dataset $\trg{\mathcal{D}}=\{x_T^{(i)} \mid x_T^{(i)} \in\mathbb{R}^{H\times W \times 3}\}$, given access to a labeled source domain dataset \resizebox{\columnwidth}{!}{$\src{\mathcal{D}}=\{(x_S^{(i)}, y_S^{(i)}) \mid x_S^{(i)} \in\mathbb{R}^{H\times W \times 3}, y_S^{(i)} \in \{0,1\}^{H\times W \times K}\}$}. We utilize a student model $g_{\theta}: \mathbb{R}^{H \times W \times 3} \rightarrow [0, 1]^{H \times W \times K} $ parameterized by $\theta$ and a teacher model $h_{\phi}: \mathbb{R}^{H \times W \times 3} \rightarrow [0, 1]^{H \times W \times K} $ parameterized by $\phi$.

\subsection{Overview}
 
Our proposed method, LangDA, leverages an EMA-updated one-stage knowledge distillation framework for online self-training (Sec. \ref{method:self_training}) to adapt across visual domains. Knowledge distillation architecture consists of two models, the student and the teacher. The student network is trained with two segmentation losses simultaneously: a supervised loss on the labeled source dataset $\src{D}$, and an unsupervised loss on the unlabeled target dataset $\trg{D}$. For the unlabeled target data, the teacher network generates pseudo-labels that are used as predictions to inform the student network on $\trg{D}$.

In addition to the supervised and unsupervised objectives for aligning image domains, we propose a Contextual Language Consistency objective in Sec. \ref{method:prompt_adapt}. This objective utilizes a context-aware caption generator (Sec. \ref{method:text_gen}) that processes the source image to generate scene text embeddings, which are then aligned with the visual features extracted from the student network via image-level alignment (Sec. \ref{method:prompt_adapt}). This alignment reinforces context awareness and domain-invariance within the student model.

Finally, the teacher network’s parameters are updated through an exponential moving average (EMA) of the student network’s weights, ensuring gradual and stable adaptation throughout training. \Cref{fig:model_arch} summarizes LangDA's architecture. The individual modules are explained in the following subsections.

\begin{figure}[t]
\centering
\begin{tcolorbox}[colframe=cyan!80, colback=white, sharp corners, boxrule=1pt, arc=5pt, rounded corners, left=7pt, right=2pt, top=0pt, bottom=0pt]
    \small 
    \small
    \begin{tcolorbox}[colframe=gray!80, colback=white, sharp corners, boxrule=1pt, arc=5pt, rounded corners, left=7pt, right=7pt, top=2pt, bottom=0pt]
      \centering
      \textbf{Caption Generation: \faImage[regular]VLM}
      \end{tcolorbox}
      \smallskip 
    \hspace{-1em}
    \faUserCircle[regular]
    \textbf{Query (Image):}
    \begin{subfigure}{0.07\textwidth}
        \includegraphics[width=\textwidth]{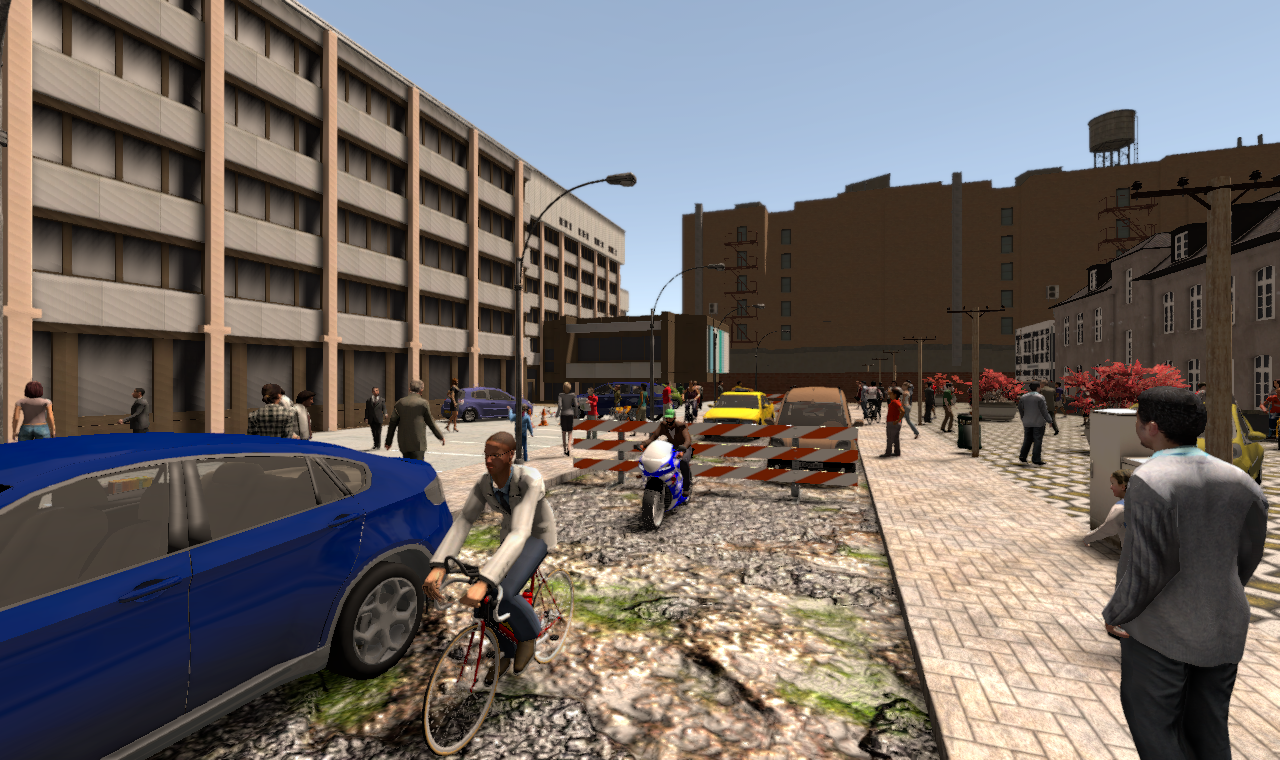}
    \end{subfigure}
    \smallskip
    
    \hspace{-1em}
    \faUserCircle[regular]
    \textbf{Query (Text):} \\
    Describe the image in detail for semantic segmentation tasks. Be sure to include the class names \textcolor{teal!80}{\{CLASS\_NAMES\}} and their pixel locations.
    \smallskip
    
    \hspace{-1em}
    \faImage[regular]
    \textbf{VLM:} \textcolor{LimeGreen!80}{\{VLM\_CAPTION\}}\\
    The image depicts a busy city street with a mix of vehicles and pedestrians. There are several \textcolor{teal!80}{cars}, including a blue \textcolor{teal!80}{car} parked on the side of the \textcolor{teal!80}{road}, and a \textcolor{teal!80}{motorcycle}. A \textcolor{teal!80}{bicycle} is also present in the scene. A \textcolor{teal!80}{person} is riding a \textcolor{teal!80}{bicycle}, while another \textcolor{teal!80}{person} is riding a \textcolor{teal!80}{motorcycle}. There are numerous people walking along the \textcolor{teal!80}{sidewalk}, some of them carrying handbags. A few pedestrians are also riding \textcolor{teal!80}{bicycles}. The street is lined with \textcolor{teal!80}{buildings}, and there is a traffic light visible in the scene. The \textcolor{teal!80}{sky} is visible in the background, adding to the urban atmosphere.
    \smallskip
    \end{tcolorbox}
    
    \vspace{-12px}
    
    \caption{\textbf{VLM Caption Generation Module.} We generate scene descriptions for source images using a VLM \cite{liu2024llava}. Class names are acquired from ground-truth labels $y_S^{(i)}$. We can see the VLM provides contextual relationships, such as "street is lined with buildings” and “numerous people walking along the sidewalk”.}
    \vspace{-15px}
    
    \label{fig:vlm-gen-ex}
\end{figure}

\begin{figure}[t]
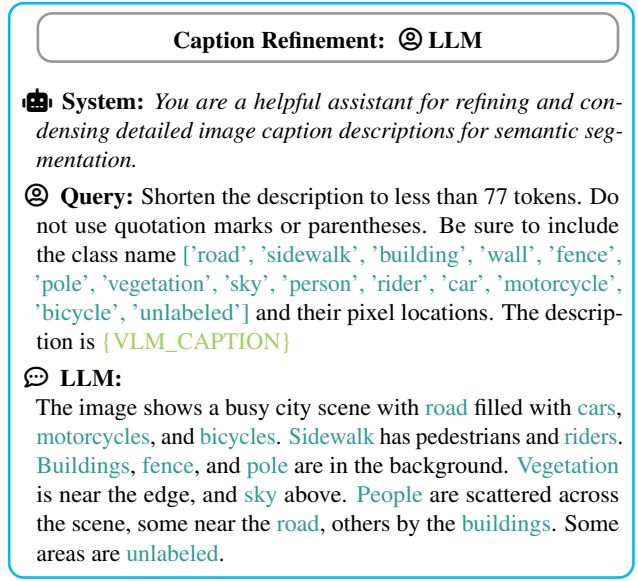

\centering
\begin{tcolorbox}[colframe=cyan!80, colback=white, sharp corners, boxrule=1pt, arc=5pt, rounded corners, left=7pt, right=2pt, top=0pt, bottom=0pt]
    \small 
    \small
    \begin{tcolorbox}[colframe=gray!80, colback=white, sharp corners, boxrule=1pt, arc=5pt, rounded corners, left=7pt, right=7pt, top=2pt, bottom=0pt]
      \centering
      \textbf{Caption Refinement: \faUserCircle[regular]LLM}
      \end{tcolorbox}
      \smallskip 
    \hspace{-1em}
    \faRobot[regular]
    \textbf{System:} \textit{You are a helpful assistant for refining and condensing detailed image caption descriptions for semantic segmentation.} 
    \smallskip
    
    \hspace{-1em}
    \faUserCircle[regular]
    \textbf{Query:} Shorten the description to less than 77 tokens. Do not use quotation marks or parentheses. Be sure to include the class name \textcolor{teal!80}{['road', 'sidewalk', 'building', 'wall', 'fence', 'pole', 'vegetation', 'sky', 'person', 'rider', 'car', 'motorcycle', 'bicycle', 'unlabeled']} and their pixel locations. The description is \textcolor{LimeGreen!80}{\{VLM\_CAPTION\}}
    \smallskip
    
    \hspace{-1em}
    \faCommentDots[regular]
    \textbf{LLM:}\\
    The image shows a busy city scene with \textcolor{teal!80}{road} filled with \textcolor{teal!80}{cars}, \textcolor{teal!80}{motorcycles}, and \textcolor{teal!80}{bicycles}. \textcolor{teal!80}{Sidewalk} has pedestrians and \textcolor{teal!80}{riders}. \textcolor{teal!80}{Buildings}, \textcolor{teal!80}{fence}, and \textcolor{teal!80}{pole} are in the background. \textcolor{teal!80}{Vegetation} is near the edge, and \textcolor{teal!80}{sky} above. \textcolor{teal!80}{People} are scattered across the scene, some near the \textcolor{teal!80}{road}, others by the \textcolor{teal!80}{buildings}. Some areas are \textcolor{teal!80}{unlabeled}.
    \smallskip
    \end{tcolorbox}
    \vspace{-12px}
    \caption{\textbf{LLM Caption Refinement Module} We summarize generated captions with an LLM. We appended a system-level prompt to inform the LLM of our semantic segmentation objective. We can see that LLM preserves spatial relationships from the VLM captions, as in “Sidewalk has pedestrians and riders”.}
    \vspace{-15px}
    \label{fig:LLaVa}
\end{figure}

\subsection{Source-target Distillation for Visual Domain}\label{method:self_training}

To adapt from a labeled source image domain $\src{\mathcal{D}}$ to an unlabeled target image domain $\trg{\mathcal{D}}$, we use single-stage online self-training (ST) \cite{tranheden2021dacs, hoyer2022daformer} to distill visual knowledge from the source to the target. This involves simultaneously training a student model $g_\theta (\cdot)$ and a teacher model $h_\phi(\cdot)$. The student segmentation network $g_\theta$ is trained on $\src{\mathcal{D}}$ using a categorical cross-entropy loss:
\begin{equation}
    \mathcal{L}_S^{(i)} = - \sum_{j=1}^{H \times W} \sum_{c=1}^C y_S^{(i,j,c)} \log g_\theta(x_S^{(i)})^{(j,c)}
\end{equation}

\noindent For target data $\trg{\mathcal{D}}$ with no labels, we generate pseudo-labels ($p_T$) via teacher network $h_\phi(\cdot)$, using the argmax of the softmax output (Eq. \eqref{eq:teacher}, $\textrm{sg}$ denotes the stopping gradient).

\begin{equation}
    p_T^{(i,j,c)} = \mathbb{I}[c = \argmax_{c'} \; \textrm{sg}(h_\phi(x_T^{(i)})^{(j,c')})]
    \label{eq:teacher}
\end{equation}
  A quality estimate for pseudo-labels is provided based on the ratio of pixels exceeding a confidence threshold $\tau$ in the softmax probability.
\begin{equation}
    q_T^{(i)} = \dfrac{1}{H \cdot W}\sum_{j=1}^{H \times W} \mathbb{I}[\max_{c'} h_\phi(x_T^{(i)})^{(j,c')} > \tau]
\end{equation}
These pseudo-labels and their confidence estimates are used to train 
$g_\theta$
  on the target domain to compute the unsupervised loss for the teacher model.
  \begin{equation}
    \mathcal{L}_T^{(i)} = - \sum_{j=1}^{H \times W} \sum_{c=1}^C q_T^{(i)} p_T^{(i,j,c)} \log g_\theta(x_T^{(i)})^{(j,c)}\,.
\end{equation}
  
  Pseudo-labels can either be generated online or offline. Following \cite{hoyer2022daformer}, we opt for online self-training due to its simplicity and single training stage, which is crucial for comparing and ablating network architectures. In online self-training, the teacher network is updated as the exponentially moving averages of the student network after each training step.
\begin{equation} \label{eq:ema_update}
    \phi_{t+1} \leftarrow \alpha \phi_t + (1 - \alpha) \theta_t
\end{equation}

\begin{table*}[h!]
\caption{\textbf{Comparison with state-of-the-art methods in UDA and Zero-shot DA.} We performed our experiments on standard adaptation benchmark Synthia $\to$ Cityscapes. Source only refers to lower bound DA baselines with no adaptation (i.e., training on source and evaluation on target). All methods' results are taken from the published paper except for those labeled with $\dagger$, which indicates the method was reproduced. Our method, when plugged into 3 existing UDA frameworks, attains state-of-the-art performance.}
\centering
\resizebox{0.8\linewidth}{!}{%
    \begin{tabular}{p{50mm} p{25mm} p{27mm} p{20mm} p{15mm} p{20mm}}
    \toprule 
    \bf Method & \bf Backbone & \bf Unlabeled Target Data & \bf Text Prompts & \bf \% mIoU $\uparrow$ \\
    \midrule
    Source only & ResNet-50 & & & \multicolumn{1}{c}{29.3} \\
    PODA$^\dagger$ \cite{fahes2023poda} & ResNet-50 & \multicolumn{1}{c}{} & \multicolumn{1}{c}{$\checkmark$} & \multicolumn{1}{c}{29.5} \\
    ULDA$^\dagger$ \cite{yang2024unified} & ResNet-50 & \multicolumn{1}{c}{} & \multicolumn{1}{c}{$\checkmark$} & \multicolumn{1}{c}{30.8} \\
    \midrule
    Source only & ResNet-101 & & & \multicolumn{1}{c}{29.4} \\
    ADVENT~\cite{vu2019advent} & ResNet-101 & \multicolumn{1}{c}{$\checkmark$} & & \multicolumn{1}{c}{41.2} \\
    CBST~\cite{zou2018unsupervised}  & ResNet-101 & \multicolumn{1}{c}{$\checkmark$} & & \multicolumn{1}{c}{42.6} \\
    DACS \cite{tranheden2021dacs}  & ResNet-101 & \multicolumn{1}{c}{$\checkmark$} & & \multicolumn{1}{c}{48.3} \\
    CorDA~\cite{wang2021domain} & ResNet-101 & \multicolumn{1}{c}{$\checkmark$} & & \multicolumn{1}{c}{55.0} \\
    ProDA~\cite{zhang2021prototypical} & ResNet-101 & \multicolumn{1}{c}{$\checkmark$} & & \multicolumn{1}{c}{55.5} \\
    \midrule
    \rowcolor[HTML]{EFEFEF} 
    DAFormer$^\dagger$ \cite{hoyer2022daformer} & MiT-B5 & \multicolumn{1}{c}{$\checkmark$} & & \multicolumn{1}{c}{61.1} \\
    \textbf{LangDA (Ours)} + DAFormer & MiT-B5 & \multicolumn{1}{c}{$\checkmark$} & \multicolumn{1}{c}{$\checkmark$} & \multicolumn{1}{c}{\cellcolor[HTML]{B7E1CD}{\textbf{62.0}}} \\
    \midrule
    \rowcolor[HTML]{EFEFEF}
    HRDA \cite{hoyer2022hrda} & MiT-B5 & \multicolumn{1}{c}{$\checkmark$} & & \multicolumn{1}{c}{65.8} \\
    \textbf{LangDA (Ours)} + HRDA & MiT-B5 & \multicolumn{1}{c}{$\checkmark$} & \multicolumn{1}{c}{$\checkmark$} & \multicolumn{1}{c}{\cellcolor[HTML]{B7E1CD}{\textbf{66.3}}} \\
    \midrule
    \rowcolor[HTML]{EFEFEF}
    MIC \cite{hoyer2023mic}  & MiT-B5 & \multicolumn{1}{c}{$\checkmark$} & & \multicolumn{1}{c}{67.3} \\
    CoPT \cite{mata2024CoPT} & MiT-B5 & \multicolumn{1}{c}{$\checkmark$} & \multicolumn{1}{c}{$\checkmark$} & \multicolumn{1}{c}{67.4} \\
    \textbf{LangDA (Ours)} + MIC & MiT-B5 & \multicolumn{1}{c}{$\checkmark$} & \multicolumn{1}{c}{$\checkmark$} & \multicolumn{1}{c}{\cellcolor[HTML]{B7E1CD}{\textbf{70.0}}} \\
    \bottomrule
    \end{tabular}
}
\label{tab:result-miou-synthia}
\end{table*}

\subsection{Context-Aware Caption Generator}\label{method:text_gen}
Capturing spatial context relations between objects is critical for dense prediction tasks \cite{hoyer2023mic}. To capture contextual information in our text prompts, we generate detailed captions describing each scene using multi-modal foundational models and utilize VLMs to generate image-level text prompts. Specifically, we choose the open-source captioning model LLaVA  \cite{liu2024llava} with GPT-4V level capabilities and is trained in a unified vision-language embedding space. LLaVA utilizes visual prompt instruction tuning and provides rich visual and linguistic context, making it ideal for our task at hand. 

We develop our captions once at model initialization and store them in a memory bank. For each image, we obtain a caption $z_S^{(i)}$ from LLaVA \cite{liu2024llava} (Fig. \ref{fig:vlm-gen-ex}), forming the caption bank \makebox{$\src{\mathcal{C}}={z_S^{(i)} \mid z_S^{(i)} \in\mathbb{R}^{\ell}}$}, where $\ell$ is the number of tokens.

To refine \makebox{$\src{\mathcal{C}}$} and obtain \makebox{$\mathcal{C}_r={z_r^{(i)} \mid z_r^{(i)} \in\mathbb{R}^{\ell}\text{, }\ell\leq77}$}, we design a system-level prompt \cite{zhang2024sys_prompt} (see Fig. \ref{fig:LLaVa}) to precisely guide the LLM’s response style and expertise, ensuring the caption is tailored for segmentation tasks. Next, we refine each $z_S^{(i)} \in \src{\mathcal{C}}$ using an LLM \cite{jiang2023mistral} to retain only the classes present in its ground truth segmentation mask $y_S^{(i)}$, as shown in Fig. \ref{fig:LLaVa}. This avoids hallucination \cite{alayrac2022flamingo,lai2024lisa}, since each source image $x_S^{(i)}$ may contain only a subset of all possible classes. Since CLIP \cite{radford2021clip} accepts a maximum token length of 77, the refinement process condenses the VLM caption from 140 tokens to approximately 70 tokens while preserving key contextual information. See Appendix \ref{sec:qual_img_desc} for an analysis of token lengths in context descriptions.

To obtain text features for image-level alignment, we pass the refined captions $\mathcal{C}_r$ to the frozen CLIP encoder $E_\text{CLIP}$ \cite{radford2021clip} to obtain the set of text feature vectors \makebox{${\mathcal{V}_{\text{CLIP}}}=\{\promptFeat^{(i)} \mid \promptFeat^{(i)} = E_\text{CLIP}({z}_r^{(i)}), \textbf{ } \promptFeat^{(i)}\in\mathbb{R}^{C}\}$}, where $C$ is the CLIP multi-modal embedding space dimension.
 
\subsection{Image-level Consistency Alignment}
\label{method:prompt_adapt}

To align source image features with generalized context-aware text features, we impose an image-level minimization objective on the distance between CLIP textual features ${\mathcal{V}_{\text{CLIP}}}$ and the source image feature $\src{\mathcal{F}}$,  \makebox{$\src{\mathcal{F}}{=}\{\src{f}^{(i)} \mid \src{f}^{(i)} {=} {g_\theta}({x}_S^{(i)}), \src{f}^{(i)} \in \mathbb{R}^{C \times H \times W}$\}}, where $C$ is the dimension of the CLIP multimodal embedding space, $H$ and $W$ are the height and width of the source image features.

To enforce image features to be in the same dimension as the CLIP embeddings, we apply attention pooling \cite{radford2021clip} and transform $\src{f}^{(i)}$ into ${f}_{\text{pool}}^{(i)}$, where ${f}_{\text{pool}}^{(i)} \in \mathbb{R}^C$. Since CLIP image encoder is trained on sparse prediction tasks such as classification, it is known to perform poorly on segmentation tasks \cite{catseg,zhou2022maskclip}. We train our own encoder on the image domain to facilitate segmentation performance. 

To align image and text features in a shared latent space, we project frozen CLIP text features and image features into the same space using a trainable multilayer perception layer (or an "adapter"). Parameter-efficient fine-tuning works have used adapters in NLP and multi-modal feature fusion without distribution shift \cite{houlsby2019parameter,song2023meta_adapter,sung2022vl_adapter}. We empirically demonstrate its effectiveness in DASS (Sec. \ref{sec:results}).

To bring text features closer to image features, we define a language consistency objective function:
 \begin{equation}
	\mathcal{L}_p^{(i)}({f}_{\text{pool}}^{(i)}, \promptFeat^{(i)}) = 1-\frac{{f}^{(i)}_{\text{pool}}\cdot\promptFeat^{(i)}}{\|{f}^{(i)}_{\text{pool}}\|\,\|\promptFeat^{(i)}\|}\,.
	\label{eqn:loss_dir}
\end{equation}

\begin{table*}[t!]
\centering
\caption{Per-class performance on Synthetic-to-Real adaptation benchmark: Synthia$\to$Cityscapes. \textbf{Bold} indicates the best score. \underline{Underline} indicates second-best score. All methods’ results are taken from publications unless otherwise indicated.}
\vspace{-8px}
\label{tab:synthia_per_class_seg_iou}
\setlength{\tabcolsep}{3pt}
\scriptsize
\resizebox{\linewidth}{!}{%
\begin{tabular}{lccccccccccccccccc}
\hline
Method & Road & S.walk & Build. & Wall & Fence & Pole & Tr.Light & Sign & Veget. &  Sky & Person & Rider & Car & Bus & M.bike & Bike &\textbf{\% mIoU $\uparrow$}\\
\hline
ADVENT~\cite{vu2019advent} & 85.6 & 42.2 & 79.7 & 8.7 & 0.4 & 25.9 & 5.4 & 8.1 & 80.4 & 84.1 & 57.9 & 23.8 & 73.3 & 36.4 & 14.2 & 33.0 & 41.2\\
DACS~\cite{tranheden2021dacs} & 80.6 & 25.1 & 81.9 & 21.5 & 2.9 & 37.2 & 22.7 & 24.0 & 83.7 & 90.8 & 67.6 & 38.3 & 82.9 & 38.9 &  28.5 & 47.6 & 48.3\\
ProDA~\cite{zhang2021prototypical} & \underline{87.8} & 45.7 & 84.6 & 37.1 & 0.6 & 44.0 & 54.6 & 37.0 & \underline{88.1} &  84.4 & 74.2 & 24.3 & 88.2 & 51.1 & 40.5 & 45.6 & 55.5\\
DAFormer~\cite{hoyer2022daformer} & 86.2 & 42.3 & 88.2 & 38.4 & \underline{8.6} & 49.9 & 55.6 & 54.1 & 86.9 & 89.3 & 73.4 & 47.1 & 87.8 & 57.3 & 53.1 & 60.2 & 61.1\\
HRDA~\cite{hoyer2022hrda} & 85.2 &47.7 & 88.8& 49.5& 4.8 & 57.2& 65.7& 60.9 & 85.3 & 92.9 & 79.4 & 52.8 & 89.0 & 64.7& 63.9& \underline{64.9} & 65.8\\
MIC~\cite{hoyer2022hrda} & 86.6& \underline{50.5}& 89.3&47.9& 7.8& 59.4&66.7 & \underline{63.4}& 87.1& \underline{94.6} & \underline{81.0} & \underline{58.9} & \underline{90.1} & 61.9 & \underline{67.1} & 64.3& 67.3\\

CoPT \cite{mata2024CoPT} & 83.4& 44.3 & \underline{90.0} & \underline{50.4} & 8.0& \underline{60.0} &\underline{67.0} & 63.0 & {87.5} & \textbf{94.8} & \textbf{81.1} & 58.6 & 89.7& \underline{66.5}& \textbf{68.9} & \textbf{65.0} & \underline{67.4}\\
\hline
\textbf{LangDA (Ours)} & \textbf{92.0} & \textbf{58.6} & \textbf{90.8} & \textbf{57.3} & \textbf{9.7} & \textbf{62.9} & \textbf{69.9} & \textbf{64.2} & \textbf{88.3} & 94.4 & 80.6 & \textbf{59.5} & \textbf{91.2} & \textbf{70.6} & 66.4 & 63.8 & \cellcolor[HTML]{B7E1CD}\textbf{70.0}\\
\hline
\end{tabular}}
\end{table*}

\begin{table*}[!t]
\centering
\caption{Per-class performance on Day-to-Night: Cityscapes$\to$DarkZurich. \textbf{Bold} indicates the best score. \underline{Underline} indicates second-best score. All methods’ results are taken from publications except for those labeled with $\dagger$, which indicates the method was retrained.}
\vspace{-8px}
\label{tab:cs2dz_per_class_seg_iou}
\setlength{\tabcolsep}{3pt}
\scriptsize
\resizebox{\linewidth}{!}{%
\begin{tabular}{lcccccccccccccccccccc}
\hline
Method & Road & S.walk & Build. & Wall & Fence & Pole & Tr.Light & Sign & Veget. & Terrain & Sky & Person & Rider & Car & Truck & Bus & Train & M.bike & Bike & \textbf{\% mIoU $\uparrow$}\\
\hline
ADVENT~\cite{vu2019advent} & 85.8 & 37.9 & 55.5 & 27.7 & 14.5 & 23.1 & 14.0 & 21.1 & 32.1 & 8.7 & 2.0 & 39.9 & 16.6 & 64.0 & 13.8 & 0.0 & 58.8 & 28.5 & 20.7 & 29.7\\
MGCDA~\cite{sakaridis2020map} & 80.3 & 49.3 & 66.2 & 7.8 & 11.0 & 41.4 & 38.9 & 39.0 & \underline{64.1} & 18.0 & 55.8 & 52.1 & 53.5 & 74.7 & 66.0 & 0.0 & 37.5 & 29.1 & 22.7 & 42.5\\
DANNet~\cite{wu2021dannet} & 90.0 & 54.0 & 74.8 & 41.0 & 21.1 & 25.0 & 26.8 & 30.2 & \textbf{72.0} & 26.2 & \textbf{84.0} & 47.0 & 33.9 & 68.2 & 19.0 & 0.3 & 66.4 & 38.3 & 23.6 & 44.3\\
DAFormer~\cite{hoyer2022daformer} & 93.5 & 65.5& 73.3 & 39.4 & 19.2 & 53.3 & \underline{44.1} & 44.0& 59.5 & 34.5& 66.6 & 53.4 & 52.7 &82.1 & 52.7 & 9.5 & 89.3 & 50.5 & 38.5 & 53.8\\
HRDA~\cite{hoyer2022hrda} & 90.4 & 56.3 & 72.0 & 39.5 & 19.5 & 57.8 & 52.7& 43.1 & 59.3 & 29.1 & \underline{70.5} &60.0& \underline{58.6} &\underline{84.0}& 75.5& \underline{11.2} & 90.5& 51.6& 40.9& 55.9\\

MIC$^\dagger$ \cite{hoyer2023mic} 
& 86.2 & 57.9 & \textbf{81.0} & \underline{51.6} & \textbf{21.4} & 61.2& 23.7 & \textbf{55.1} & 57.3 & 42.0& 59.0 & \underline{62.2} & 55.4& 65.2& \underline{78.6} & 5.22 & 90.0 & \underline{53.4} & 42.3 & 55.2 \\
CoPT \cite{mata2024CoPT}& \textbf{92.7} & \underline{66.1} & \underline{80.0} & 49.0 & 19.3 & \textbf{63.2} & \textbf{51.7} &  52.0 & 50.9 & \underline{43.1} & 55.9 & 61.7 & 56.5 & 59.6 & \textbf{79.4} & 2.96 & \underline{90.8}  &50.0 & \underline{42.5} & \underline{56.2}\\
\hline
\textbf{LangDA (Ours)} & \underline{92.0} & \textbf{68.4} & 79.6 &\textbf{53.4} & \underline{19.9} & \underline{61.7} & 32.1 & \underline{54.7}  &44.7
& \textbf{44.0} & 50.9 &\textbf{62.7}& \textbf{60.0} & \textbf{84.1} & 78.2 & \textbf{15.4} & \textbf{92.1 }&\textbf{58.7}& \textbf{43.8}& \cellcolor[HTML]{B7E1CD}\textbf{57.6}\\
\hline
\end{tabular}}
\end{table*}

\begin{table*}[!t]
\centering
\caption{Per-class performance on Clear-to-Adverse-Weather: Cityscapes$\to$ACDC. \textbf{Bold} indicates the best score. \underline{Underline} indicates second-best score. All methods’ results are taken from publications unless otherwise indicated.}
\vspace{-8px}
\label{tab:cs2acdc_per_class_seg_iou}
\setlength{\tabcolsep}{3pt}
\scriptsize
\resizebox{\linewidth}{!}{%
\begin{tabular}{lcccccccccccccccccccc}
\hline
Method & Road & S.walk & Build. & Wall & Fence & Pole & Tr.Light & Sign & Veget. & Terrain & Sky & Person & Rider & Car & Truck & Bus & Train & M.bike & Bike & \% mIoU $\uparrow$\\
\hline
ADVENT~\cite{vu2019advent} & 72.9 & 14.3 & 40.5 & 16.6 & 21.2 & 9.3 & 17.4 & 21.2 & 63.8 & 23.8 & 18.3 & 32.6 & 19.5 & 69.5 & 36.2 & 34.5 & 46.2 & 26.9 & 36.1 & 32.7\\
MGCDA~\cite{sakaridis2020map} & 73.4 & 28.7 & 69.9 & 19.3 & 26.3 & 36.8 & \textbf{53.0} & 53.3 & \textbf{75.4} & 32.0 & \underline{84.6} & 51.0 & 26.1 & 77.6 & 43.2 & 45.9 & 53.9 & 32.7 & 41.5 & 48.7\\
DANNet~\cite{wu2021dannet} & \underline{84.3} & 54.2 & 77.6 & 38.0 & 30.0 & 18.9 & 41.6 & 35.2 & 71.3 & 39.4 & \textbf{86.6}& 48.7 & 29.2 & 76.2 & 41.6 & 43.0 & 58.6 & 32.6 & 43.9 & 50.0\\
DAFormer~\cite{hoyer2022daformer} & 58.4 & 51.3 & \underline{84.0} & 42.7 & 35.1 & 50.7 & 30.0 & 57.0 & \underline{74.8} & 52.8 & 51.3 & 58.3 & 32.6 & 82.7 & 58.3 & 54.9 & \underline{82.4} & 44.1 & 50.7 & 55.4\\
CoPT \cite{mata2024CoPT} & 49.1 & \textbf{70.3} & 83.6 & \textbf{59.4} & \textbf{42.4} & \textbf{58.5} & \underline{48.3} & \textbf{67.2} & 73.5 & \textbf{60.7} & 45.0 & \underline{69.3} & \underline{45.2} & \underline{83.4} & \textbf{76.3} & \underline{74.5} & \textbf{88.2} & \underline{54.4} & \underline{61.4} & \underline{63.7}\\
\hline
\textbf{LangDA} (Ours)  & \textbf{84.7}& \underline{63.0} & \textbf{88.2} & \underline{56.4} & \underline{40.9} & \underline{57.5} & 42.9 & \underline{64.7} & \textbf{75.4} & \underline{59.4} & 82.8 & \textbf{69.8} & \textbf{46.6} & \textbf{89.1} & \underline{74.8} & \textbf{83.2} & \textbf{88.2} & \textbf{55.9} & \textbf{61.8} &\cellcolor[HTML]{B7E1CD}\textbf{67.6}\\
\hline
\multicolumn{18}{l}{}
\vspace{-18px}
\end{tabular}}
\end{table*}

To the best of our knowledge, we are the first work to employ this exact language consistency loss in DASS. Our loss leverages the cosine similarity distance metric, used by CLIP \cite{radford2021clip} for aligning multimodal embeddings, to explicitly align source image features with text embeddings. The language consistency objective guides the source image features toward the text embedding. Additionally, the global target image features are implicitly steered towards the text feature space through the EMA model update in Eq.\eqref{eq:ema_update}, further reinforcing the learned multimodal consistency.
The overall UDA loss $\mathcal{L}$ is a minimization problem of the weighted sum of the supervised loss, unsupervised loss and language consistency loss $\mathcal{L} = \mathcal{L}_S + \mathcal{L}_T + \lambda_\text{p} \mathcal{L}_\text{p}$.

\section{Experiments}
\label{sec:experiments}

\smallskip\noindent\textbf{Datasets}
Following standard practices in DASS, we explore three self-driving adaptation scenarios: synthetic-to-real, clear-to-adverse-weather, day-to-night; using Synthia $\to$ Cityscapes, Cityscapes $\to$ ACDC, and Cityscapes $\to$ DarkZurich, respectively. Synthia \cite{Ros_2016_CVPR_synthia} serves as the synthetic source, containing 9,400 images at $1280\times760$ resolution. Real dataset benchmarks include Cityscapes \cite{cordts2016cityscapes} with 2,975 training and 500 validation images at 2048 $\times$ 1024 resolution for clear weather, DarkZurich \cite{sakaridis2019darkzurich} with 2,416 training and 151 test images for nighttime, and ACDC \cite{sakaridis2021acdc} with 1,600 training and 2,000 test images for adverse weather.

\smallskip\noindent\textbf{Implementation details}
To evaluate LangDA's robustness, we combine LangDA with several vision-only DASS methods (DAFormer \cite{hoyer2022daformer}, HRDA \cite{hoyer2022hrda}, and MIC \cite{hoyer2023mic}) and evaluate on Synthia$\to$Cityscapes. For main experiments, we build on MIC, the state-of-the-art vision-only DASS method. LangDA uses LLaVA \cite{liu2024llava} to generate spatially aware scene captions, refined with Mistral-Large-2 \cite{jiang2023mistral}, which supports a 128k token context window and has an Apache 2.0 license. Refined captions are encoded using a frozen ViT-B/16 CLIP text encoder \cite{radford2021clip}. All backbones are initialized with ImageNet \cite{russakovsky2015imagenet} pretrained. For the default UDA setting, we follow the training scheme used by HRDA \cite{hoyer2022hrda}. Specifically, we utilize the AdamW \cite{adamw} with a learning rate of $6 {\times} 10^{-5}$ for the encoder and $6 {\times} 10^{-4}$ for the decoder, a batch size of 2, linear learning rate warmup, a loss weight $\lambda{=}1$, an EMA factor $\alpha{=}0.999$, quality threshold $\tau=0.968$, DACS \cite{tranheden2021dacs} data augmentation, Rare Class Sampling \cite{hoyer2022daformer}, and ImageNet Feature Distance \cite{hoyer2022daformer}. All experiments were conducted on one NVIDIA RTX A6000 GPU.
All our implementations were carried out using the MMSegmentation framework \cite{mmseg2020}. 

\section{Results}
\label{sec:results}

\subsection{Main Results}

\subsubsection{Quantitative Results on Synthia \texorpdfstring{$to$} CS.}
To assess the proposed method LangDA, we quantitatively compare LangDA with existing methods on synthetic-to-real adaptation benchmark, Synthia $\to$ Cityscapes. We evaluate our method using the standard semantic segmentation metric, mean Jaccard Index (also termed mean Intersection over Union, or, mIoU). We report the results in \Cref{tab:result-miou-synthia}, where it is evident that LangDA significantly outperforms the existing state-of-the-art by 2.6\%. Further, we can see that LangDA, when combined with current methods, consistently improves their performance across the board, ranging from 0.9\% to 2.7\%. Moreover, LangDA does not only benefit transformer architectures like DAFormer but also boosts performance when combined with complex approaches with additional regularization and are multi-resolution (HRDA, MIC). Our results demonstrate the importance of explicitly inducing context awareness via text. 

\begin{table}[b]
\centering
\caption{Performance under different weather conditions on Clear-to-Adverse-Weather benchmark: Cityscapes$\to$ACDC}
\vspace{-8px}
\label{tab:acdc-condition}
\setlength{\tabcolsep}{3pt}
\scriptsize
\resizebox{0.8\columnwidth}{!}{%
\begin{tabular}{lccccc}
\hline
Method & Rain & Snow & Fog & Night & \textbf{All}\\
\hline
MIC \cite{hoyer2023mic} &
\textbf{74.2} & 60.7 & 56.8 & \textbf{57.6} & 63.3 \\
CoPT \cite{mata2024CoPT} & 74.0 & 60.6 & 60.4 & 56.8 & 63.7\\
\hline
\textbf{LangDA (Ours)} & 73.3 & \cellcolor[HTML]{B7E1CD}\textbf{68.3}& \cellcolor[HTML]{B7E1CD}\textbf{68.9} &55.7 & \cellcolor[HTML]{B7E1CD}\textbf{67.6}\\
\hline
\end{tabular}%
}
\vspace{-10pt}
\end{table}

\subsubsection{Per-class IoU on Synthia \texorpdfstring{$\to$} Cityscapes} 
\begin{figure*}[htbp]
    \centering
    \begin{minipage}[t]{0.24\textwidth}
        \text{Image}
        \centering
        \begin{tikzpicture}
            \node[anchor=south west, inner sep=0] (image) at (0,0) { \includegraphics[width=\linewidth]{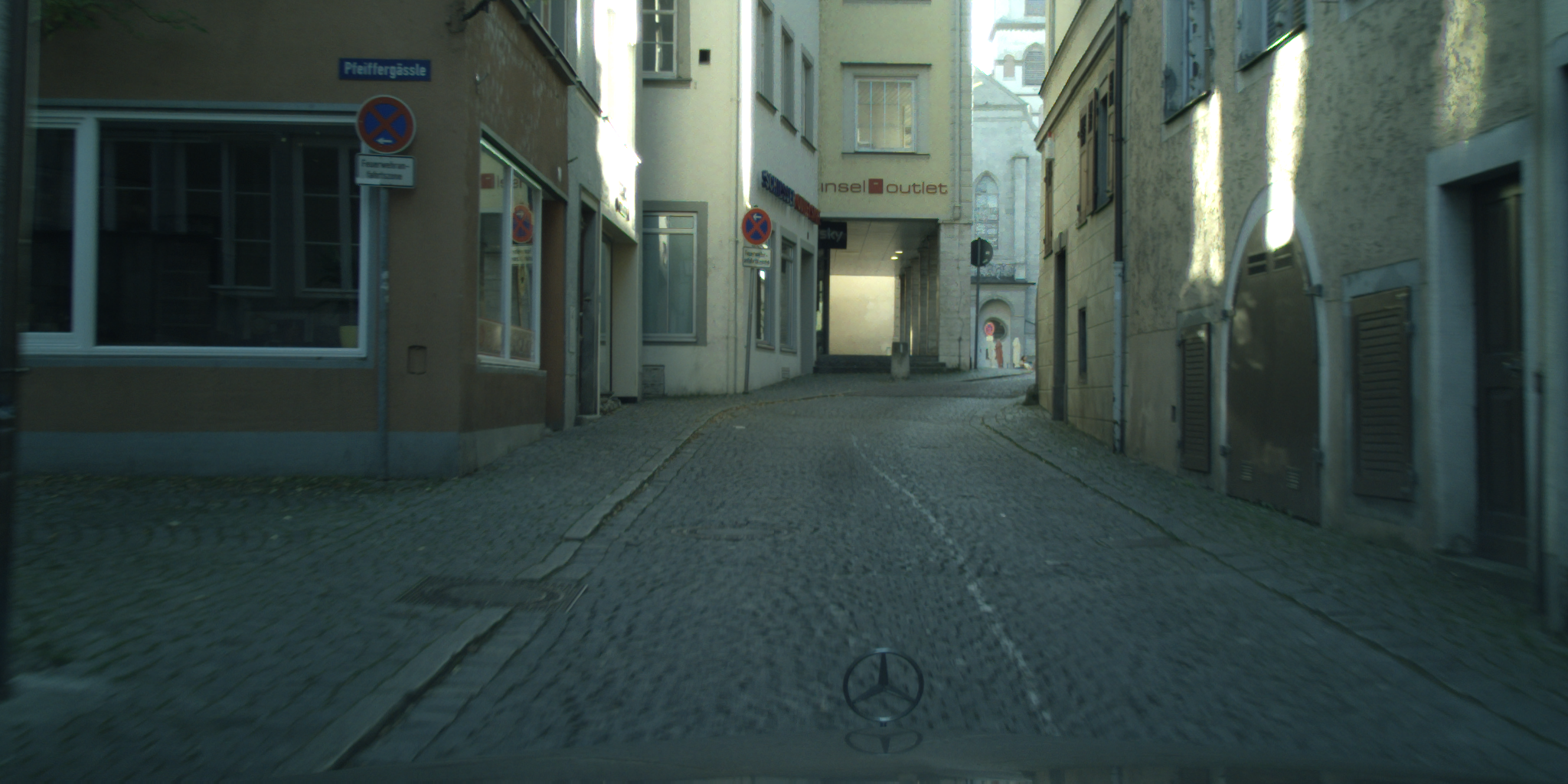} };
            \begin{scope}[x={(image.south east)}, y={(image.north west)}]
                \draw[white, dashed] (0.4, 0.2) rectangle (0.75, 0.4);  
            \end{scope}
        \end{tikzpicture}
        \vspace{0.5ex}

        \begin{tikzpicture}
            \node[anchor=south west, inner sep=0] (image) at (0,0) { \includegraphics[width=\linewidth]{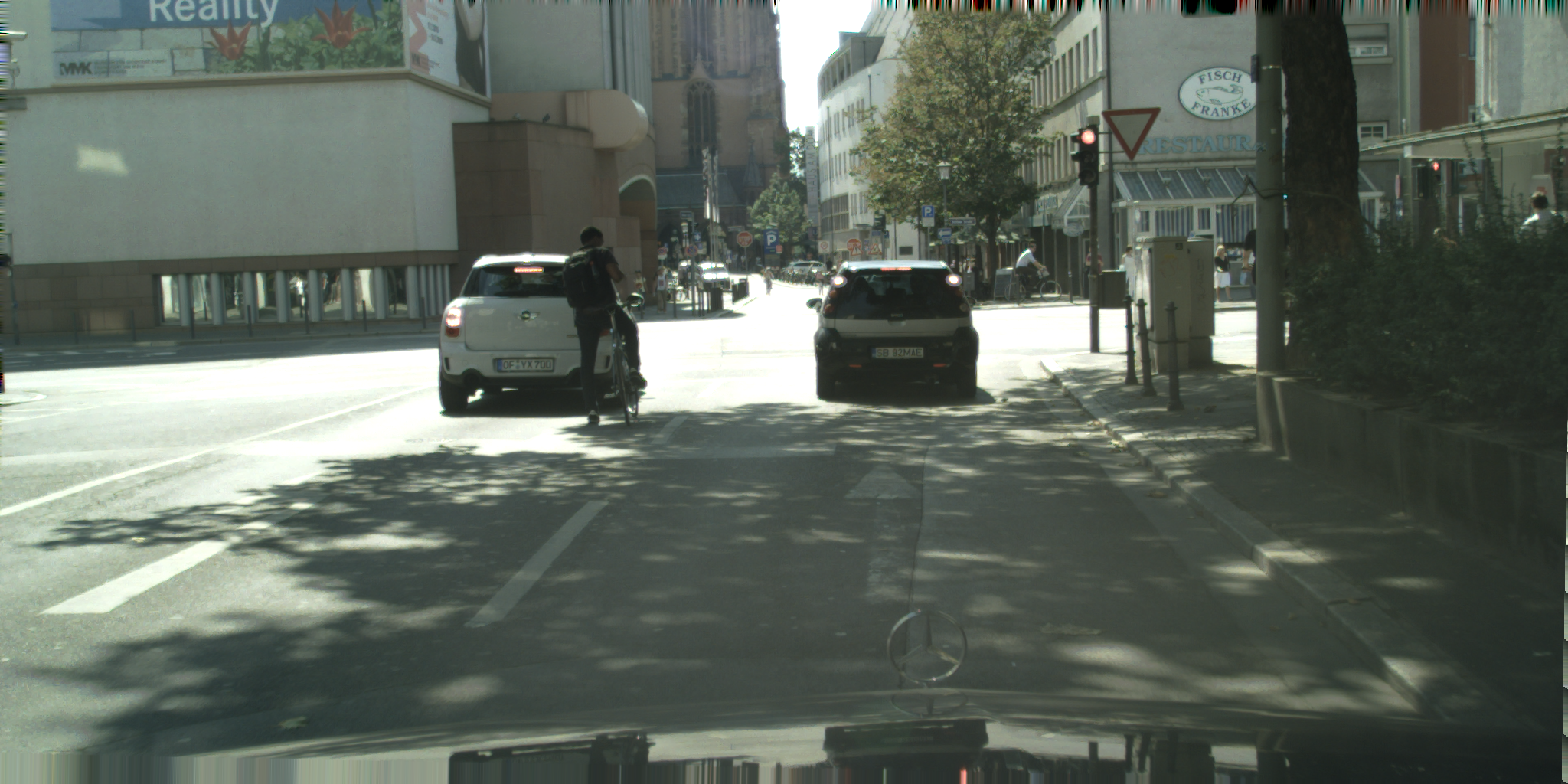} };
            \begin{scope}[x={(image.south east)}, y={(image.north west)}]
                \draw[white, dashed] (0.85, 0.1) rectangle (0.97, 0.4);
            \end{scope}
        \end{tikzpicture}
        \vspace{0.5ex}
        
        \begin{tikzpicture}
            \node[anchor=south west, inner sep=0] (image) at (0,0) { \includegraphics[width=\linewidth]{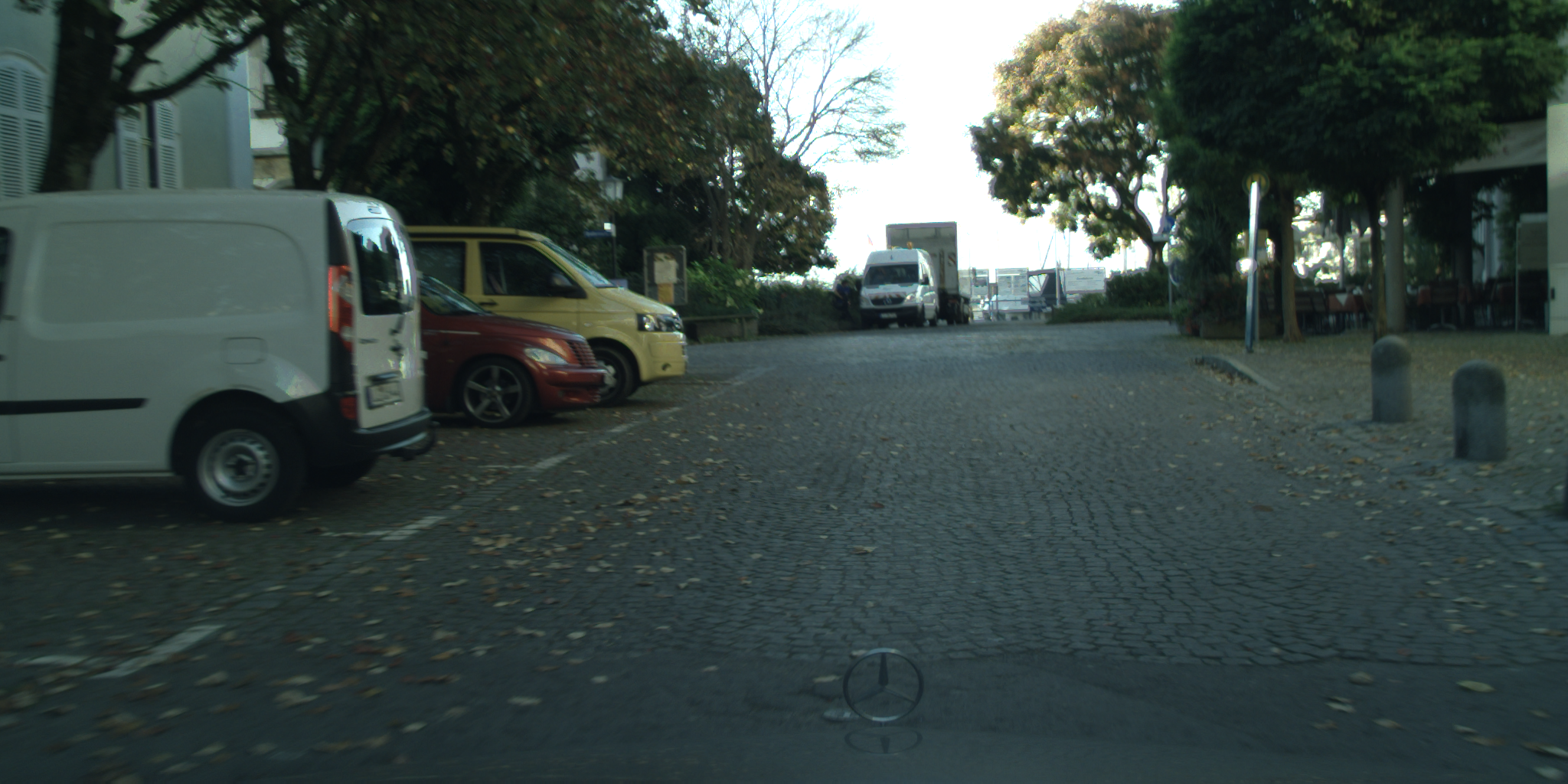} };
            \begin{scope}[x={(image.south east)}, y={(image.north west)}]
                \draw[white, dashed] (0.2, 0.2) rectangle (0.95, 0.6);  
                \end{scope}
        \end{tikzpicture}
        \vspace{0.5ex}
        
        \begin{tikzpicture}
            \node[anchor=south west, inner sep=0] (image) at (0,0) { \includegraphics[width=\linewidth]{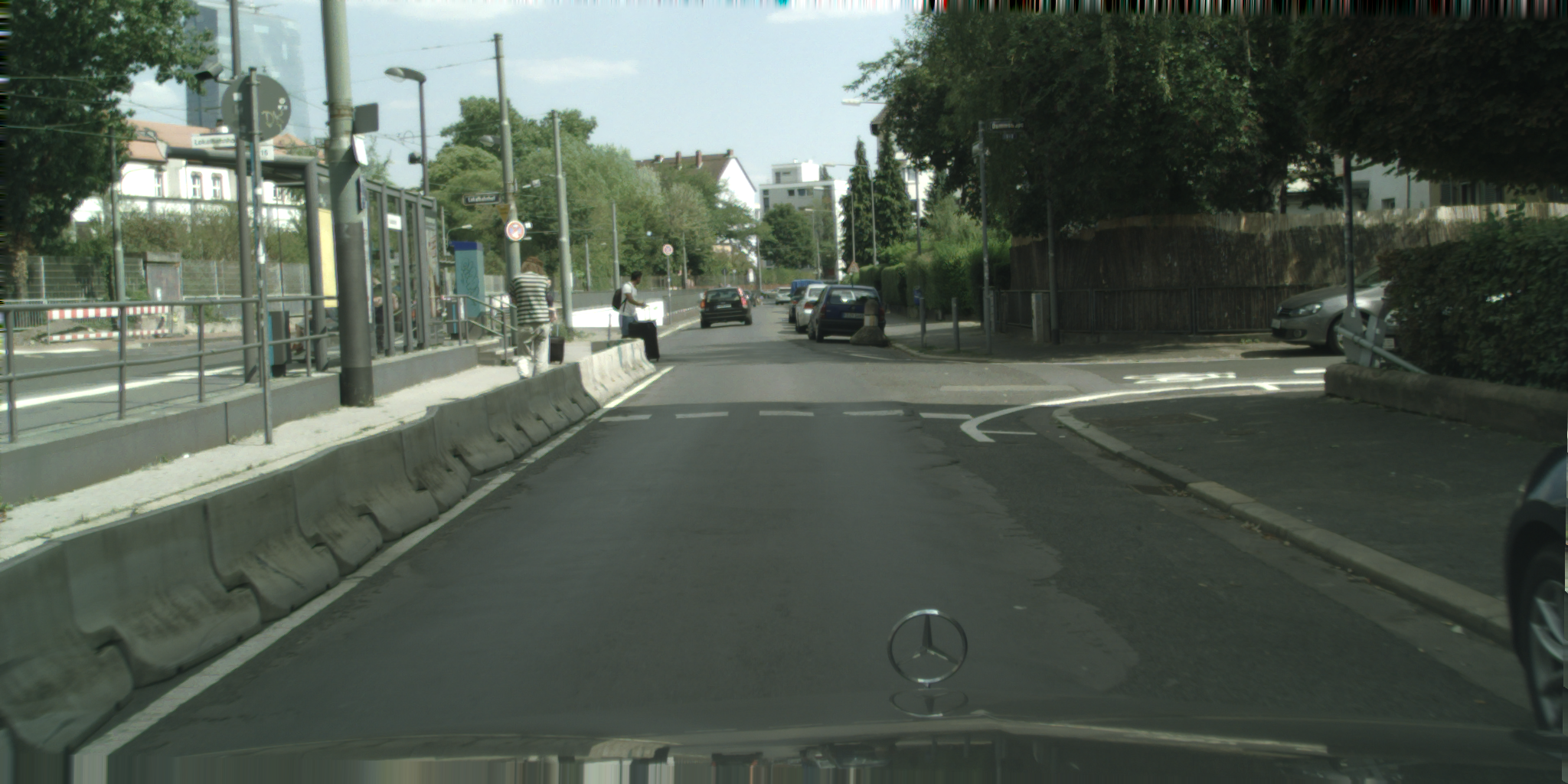} };
            \begin{scope}[x={(image.south east)}, y={(image.north west)}]
                \draw[white, dashed] (0.05, 0.25) rectangle (0.3, 0.6);  
            \end{scope}
        \end{tikzpicture}
    \end{minipage}%
    \hfill
    \begin{minipage}[t]{0.24\textwidth}
        \text{MIC \cite{hoyer2023mic}}
        \centering
        \begin{tikzpicture}
            \node[anchor=south west, inner sep=0] (image) at (0,0) { \includegraphics[width=\linewidth]{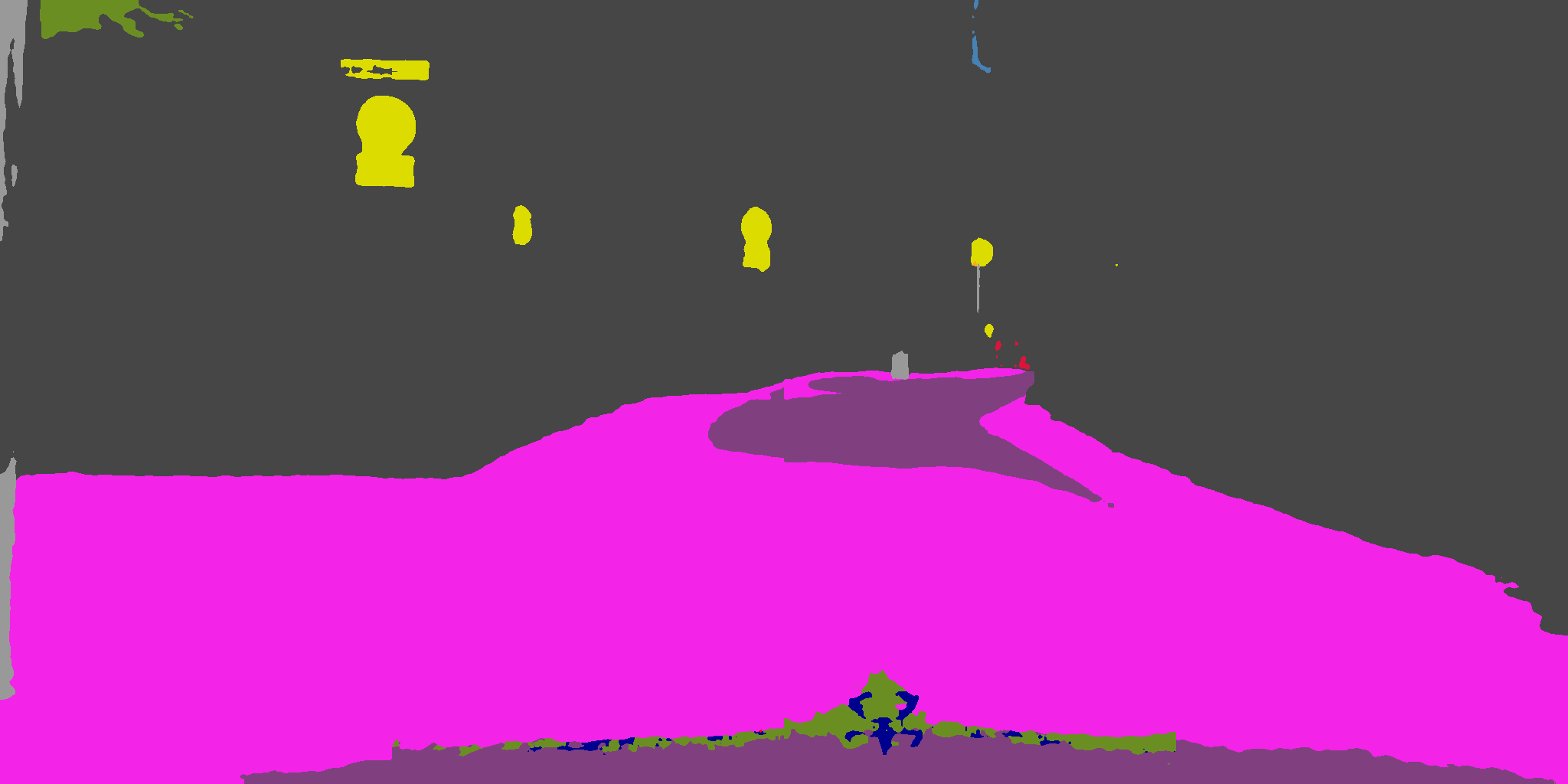} };
            \begin{scope}[x={(image.south east)}, y={(image.north west)}]
                \draw[white, dashed] (0.4, 0.2) rectangle (0.75, 0.4);  
            \end{scope}
        \end{tikzpicture}
        \vspace{0.5ex}

        \begin{tikzpicture}
            \node[anchor=south west, inner sep=0] (image) at (0,0) { \includegraphics[width=\linewidth]{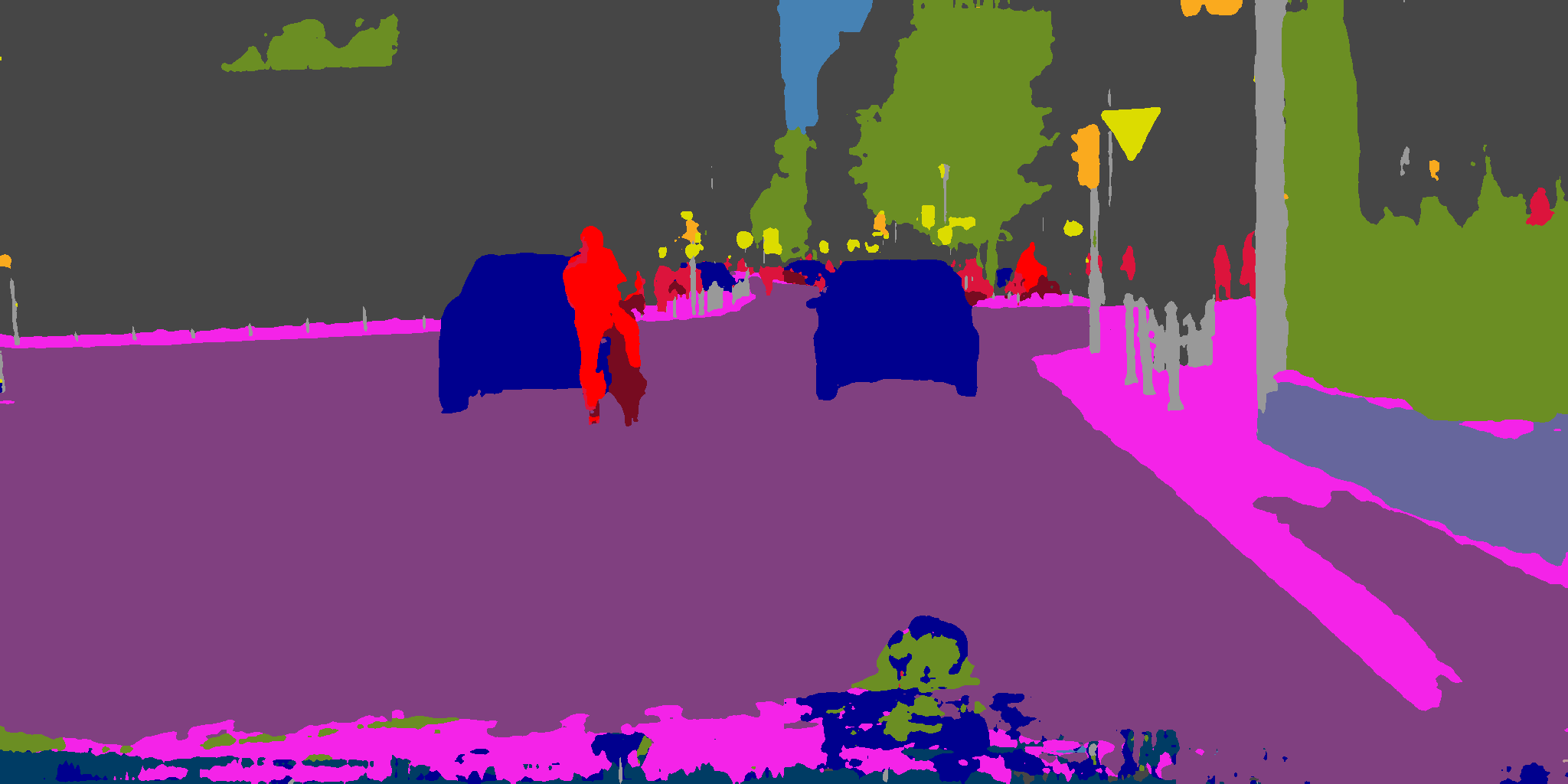} };
            \begin{scope}[x={(image.south east)}, y={(image.north west)}]
                \draw[white, dashed] (0.85, 0.1) rectangle (0.97, 0.4);
            \end{scope}
        \end{tikzpicture}
        \vspace{0.5ex}
        
        \begin{tikzpicture}
            \node[anchor=south west, inner sep=0] (image) at (0,0) { \includegraphics[width=\linewidth]{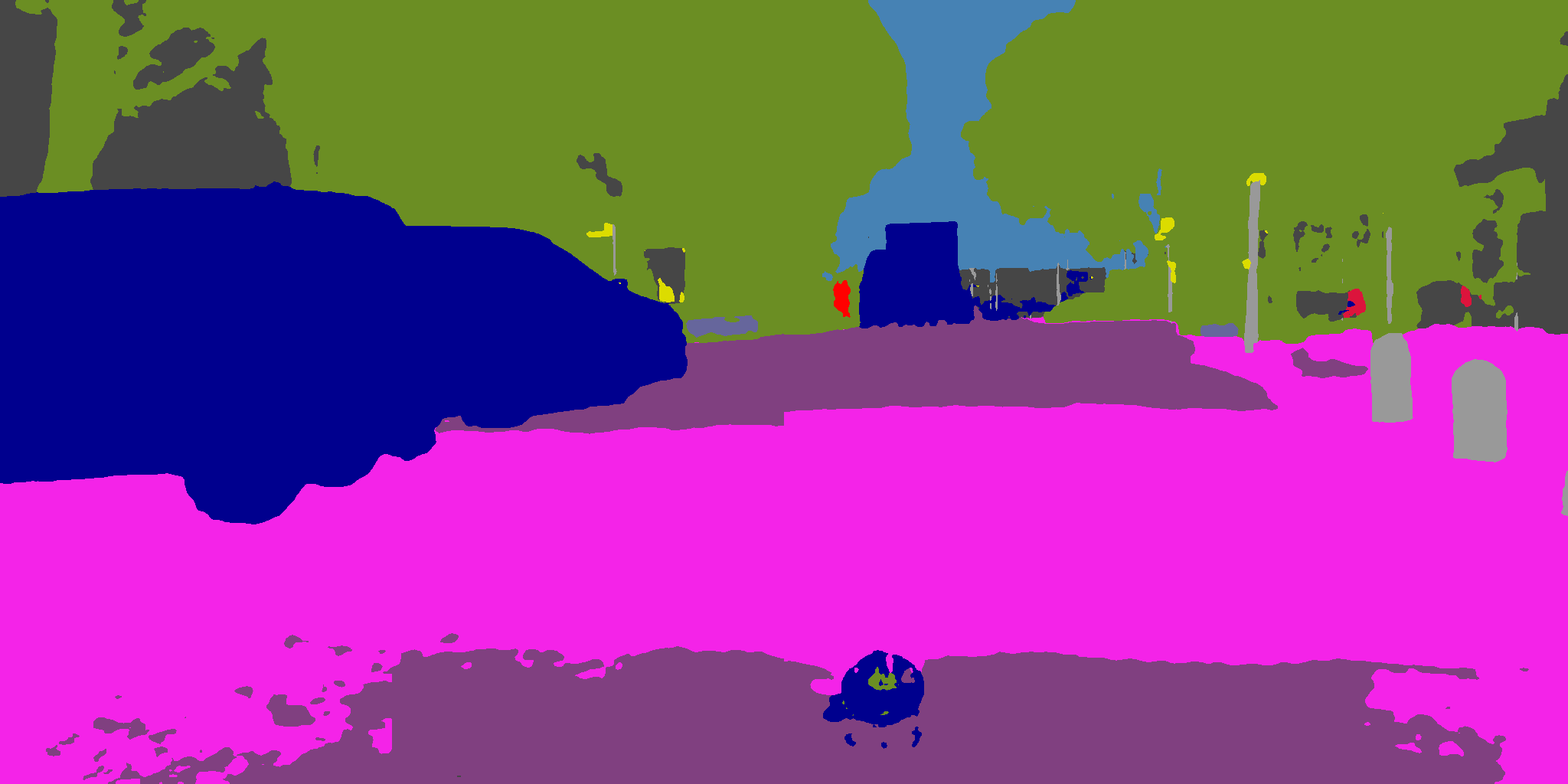} };
            \begin{scope}[x={(image.south east)}, y={(image.north west)}]
                \draw[white, dashed] (0.2, 0.2) rectangle (0.95, 0.6);  
                \end{scope}
        \end{tikzpicture}
        \vspace{0.5ex}
        
        \begin{tikzpicture}
            \node[anchor=south west, inner sep=0] (image) at (0,0) { \includegraphics[width=\linewidth]{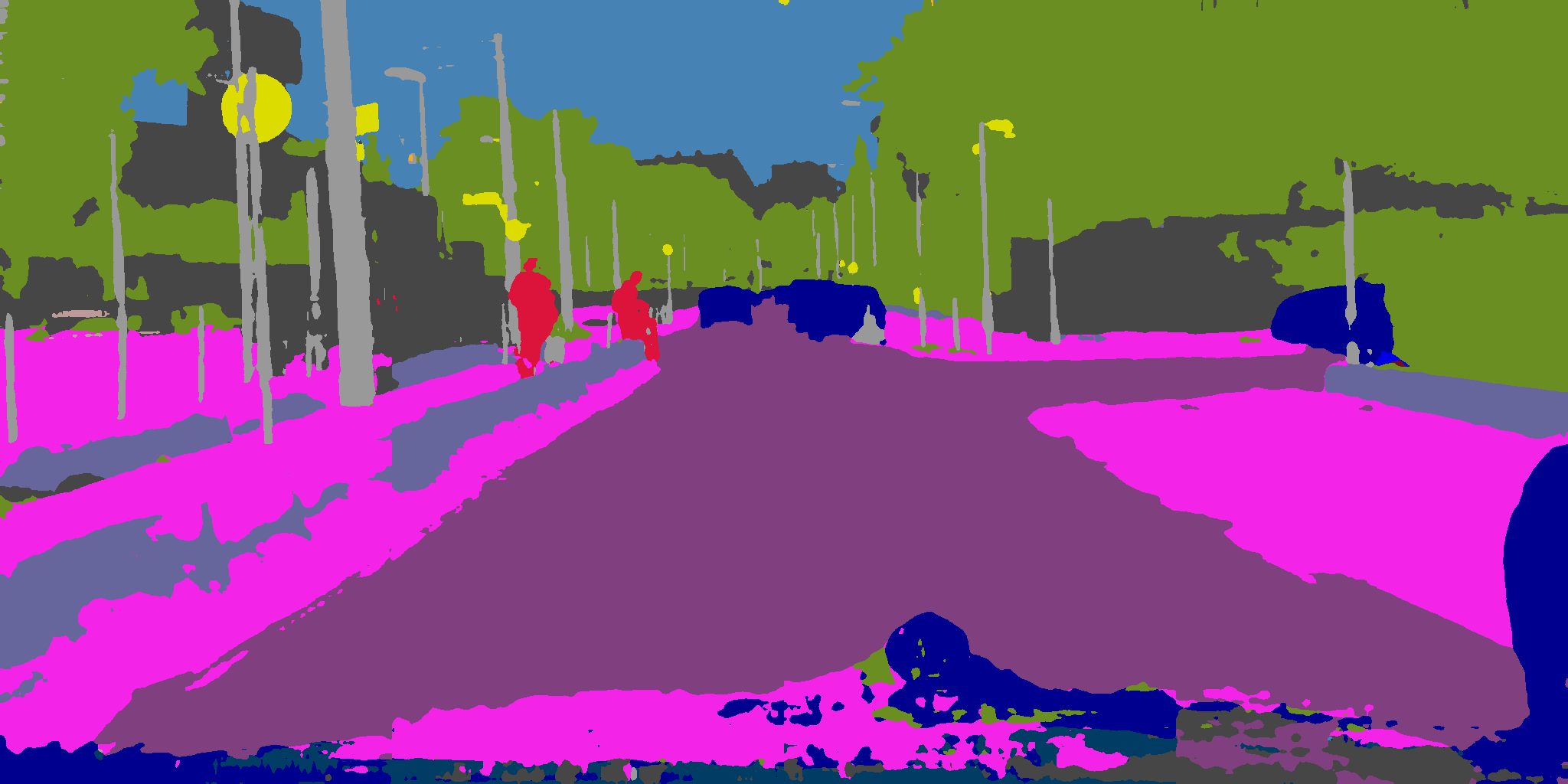} };
            \begin{scope}[x={(image.south east)}, y={(image.north west)}]
                \draw[white, dashed] (0.05, 0.25) rectangle (0.3, 0.6);  
            \end{scope}
        \end{tikzpicture}
    \end{minipage}
    \hfill
    \begin{minipage}[t]{0.24\textwidth}
        \text{LangDA (Ours)}
        \centering
        \begin{tikzpicture}
            \node[anchor=south west, inner sep=0] (image) at (0,0) { \includegraphics[width=\linewidth]{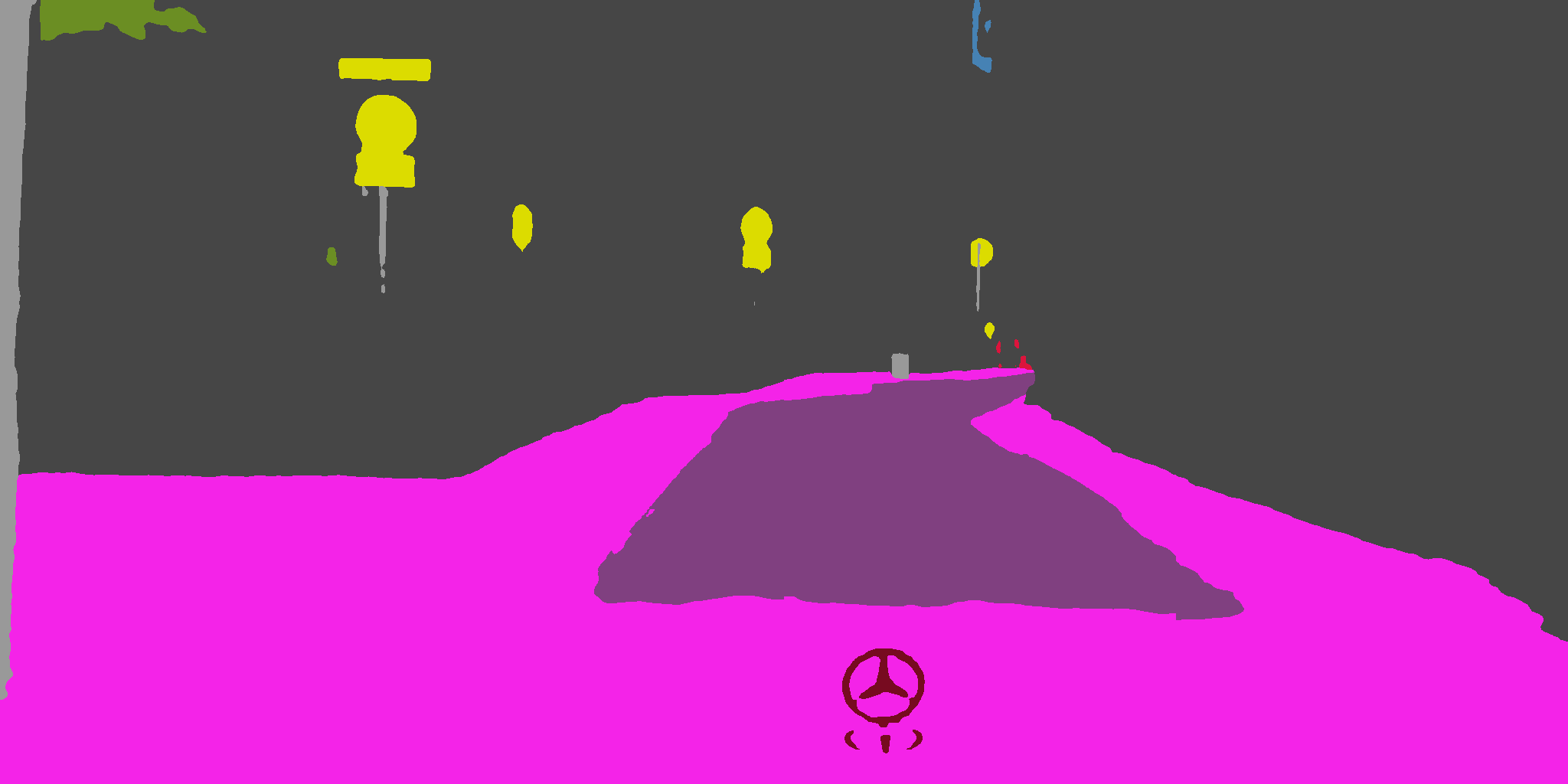} };
            \begin{scope}[x={(image.south east)}, y={(image.north west)}]
                \draw[white, dashed] (0.4, 0.2) rectangle (0.75, 0.4);  
            \end{scope}
        \end{tikzpicture}
        \vspace{0.5ex}

        \begin{tikzpicture}
            \node[anchor=south west, inner sep=0] (image) at (0,0) { \includegraphics[width=\linewidth]{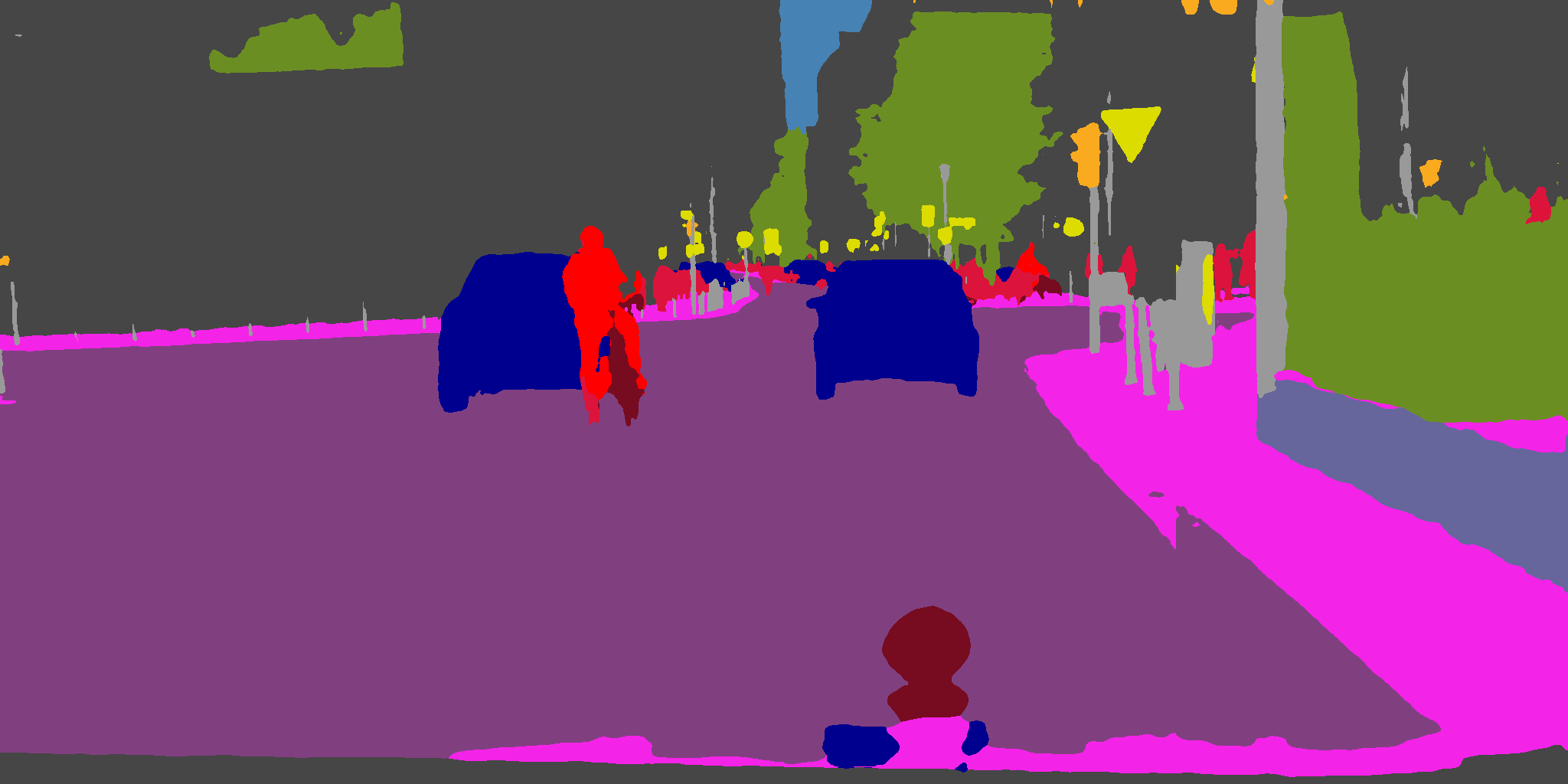} };
            \begin{scope}[x={(image.south east)}, y={(image.north west)}]
                \draw[white, dashed] (0.85, 0.1) rectangle (0.97, 0.4);  
            \end{scope}
            \end{tikzpicture}
        \vspace{0.5ex}

        \begin{tikzpicture}
            \node[anchor=south west, inner sep=0] (image) at (0,0) { \includegraphics[width=\linewidth]{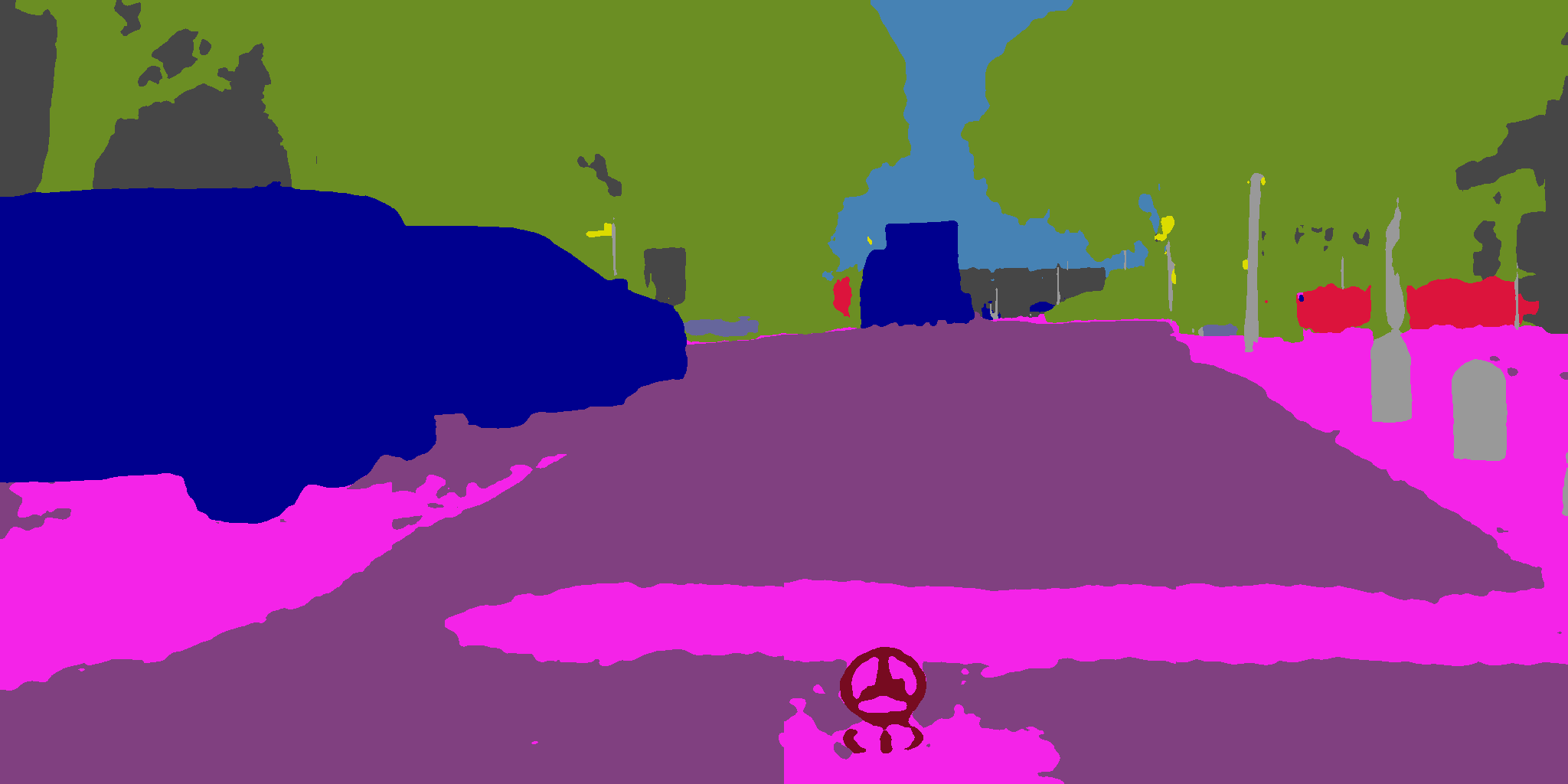} };
            \begin{scope}[x={(image.south east)}, y={(image.north west)}]
                \draw[white, dashed] (0.2, 0.2) rectangle (0.95, 0.6);  
                \end{scope}
            \end{tikzpicture}
        \vspace{0.5ex}

        \begin{tikzpicture}
            \node[anchor=south west, inner sep=0] (image) at (0,0) { \includegraphics[width=\linewidth]{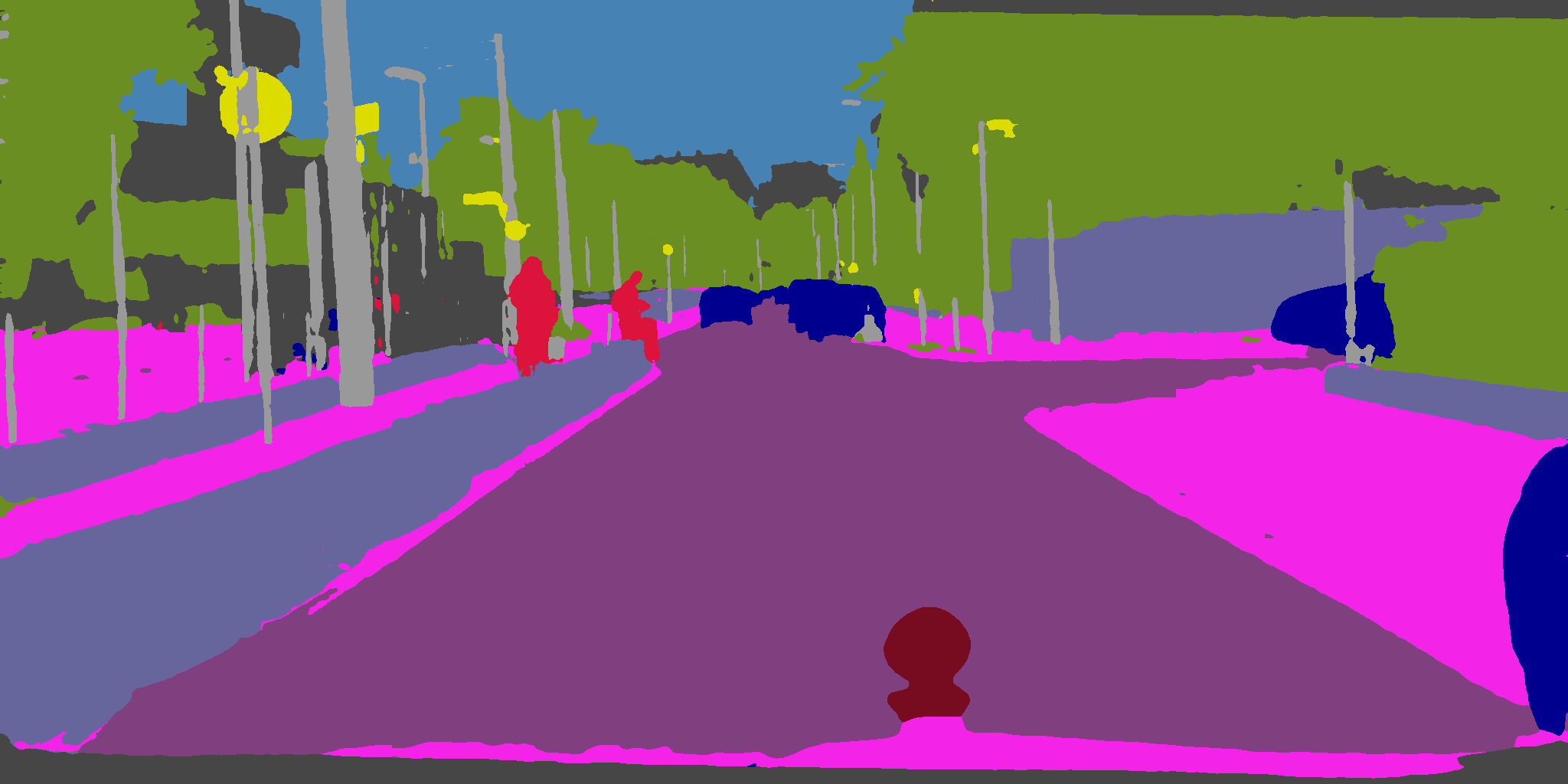} };
            \begin{scope}[x={(image.south east)}, y={(image.north west)}]
                \draw[white, dashed] (0.05, 0.25) rectangle (0.3, 0.6);  
            \end{scope}
            \end{tikzpicture}
    \end{minipage}
    \hfill
    \begin{minipage}[t]{0.24\textwidth}
        \text{Ground Truth}
        \centering
        \begin{tikzpicture}
            \node[anchor=south west, inner sep=0] (image) at (0,0) { \includegraphics[width=\linewidth]{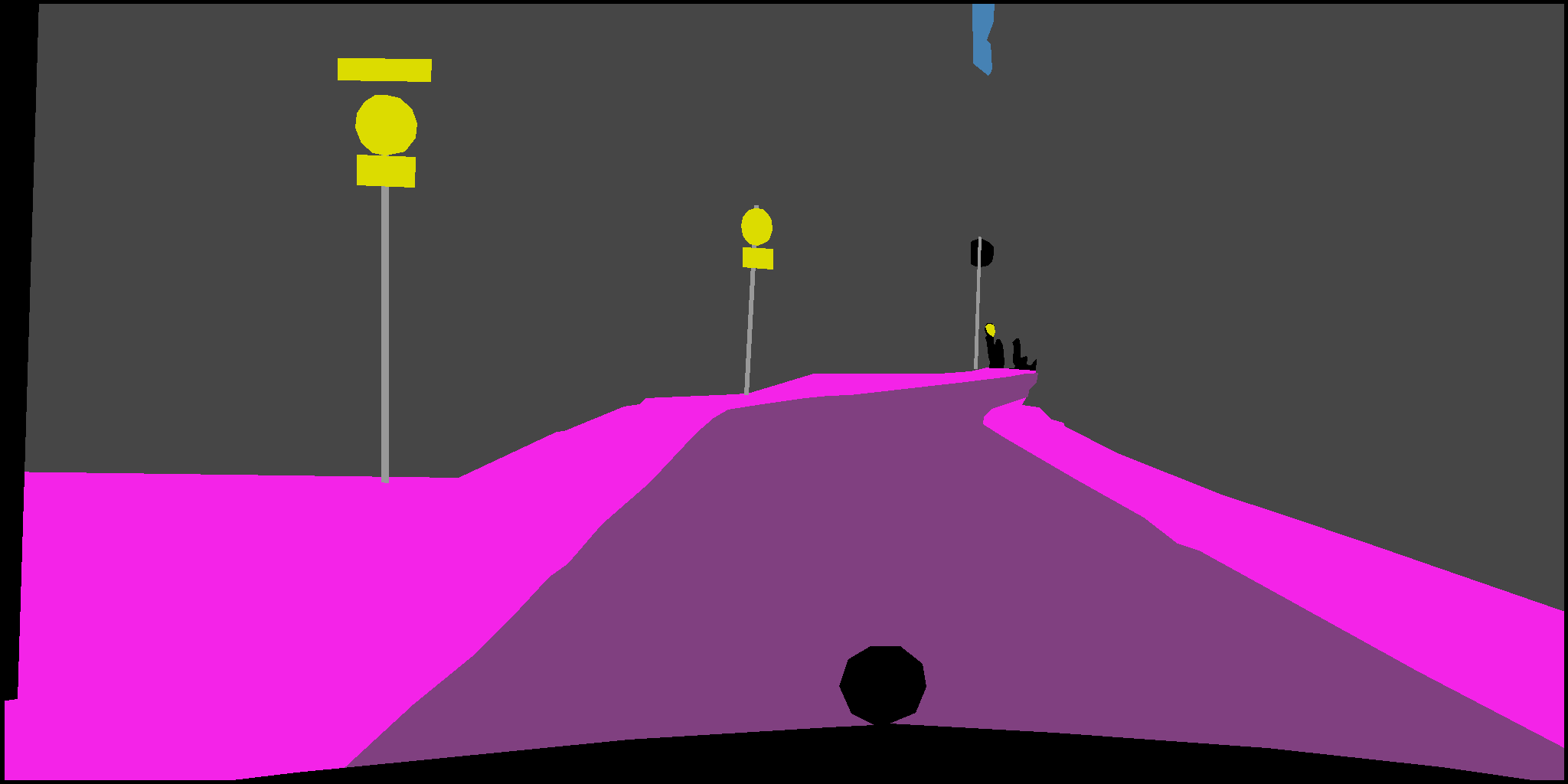} };
            \begin{scope}[x={(image.south east)}, y={(image.north west)}]
                \draw[white, dashed] (0.4, 0.2) rectangle (0.75, 0.4);  
            \end{scope}
        \end{tikzpicture}
        \vspace{0.5ex}

        \begin{tikzpicture}
            \node[anchor=south west, inner sep=0] (image) at (0,0) { \includegraphics[width=\linewidth]{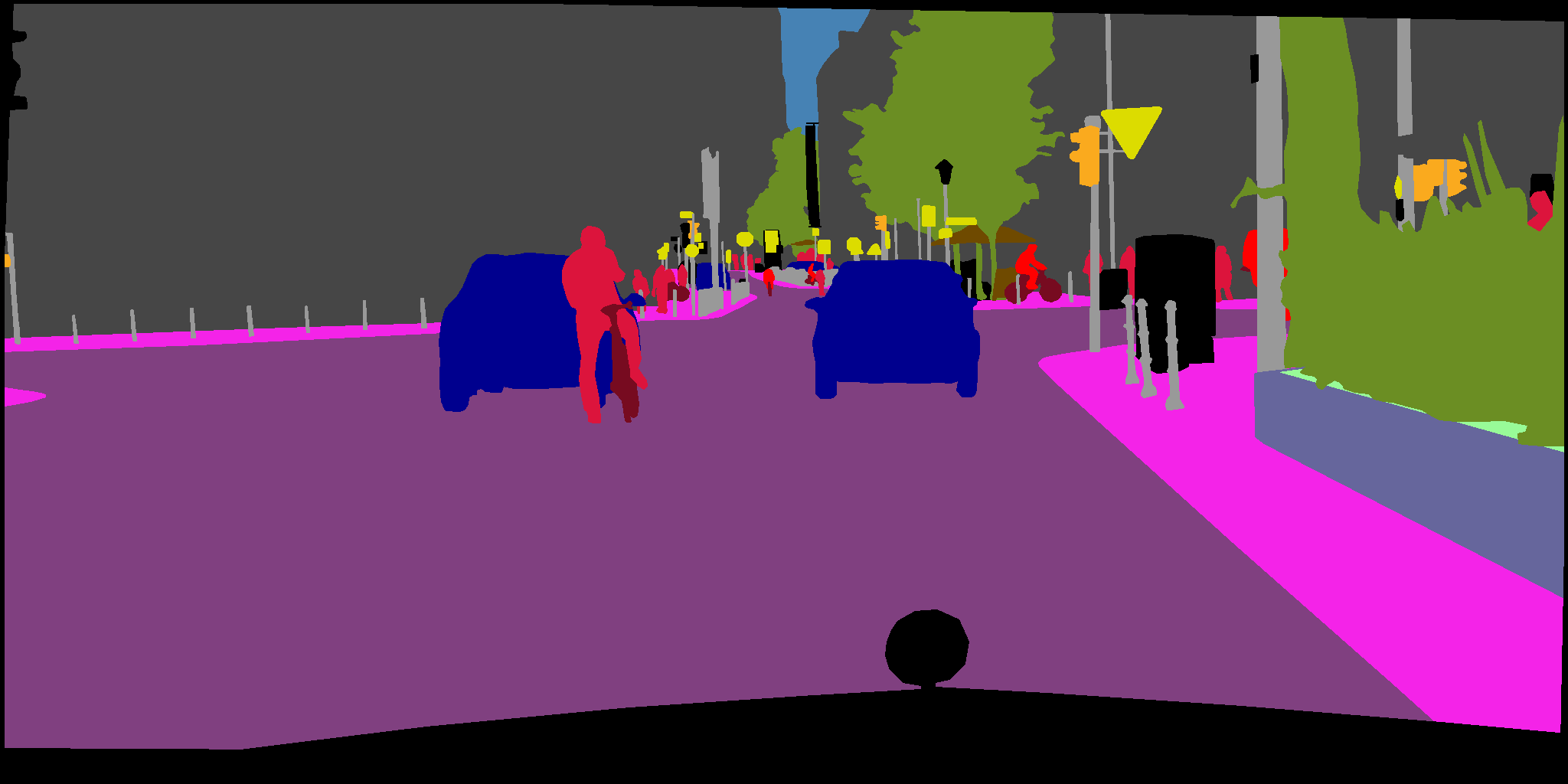} };
            \begin{scope}[x={(image.south east)}, y={(image.north west)}]
                \draw[white, dashed] (0.85, 0.1) rectangle (0.97, 0.4);  
            \end{scope}
            \end{tikzpicture}
        \vspace{0.5ex}
        
        \begin{tikzpicture}
            \node[anchor=south west, inner sep=0] (image) at (0,0) { \includegraphics[width=\linewidth]{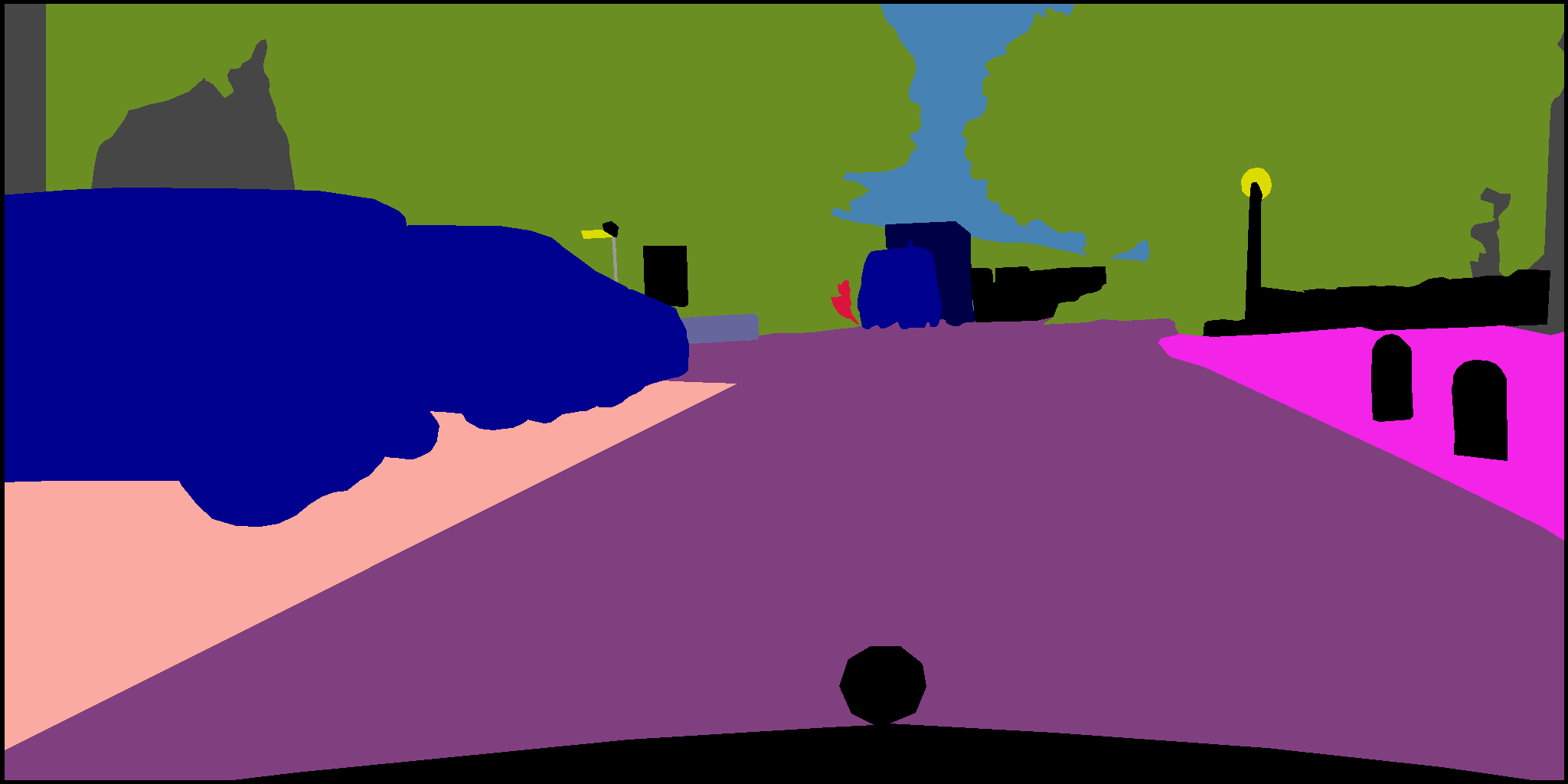} };
            \begin{scope}[x={(image.south east)}, y={(image.north west)}]
                \draw[white, dashed] (0.2, 0.2) rectangle (0.95, 0.6);  
                \end{scope}
            \end{tikzpicture}
        \vspace{0.5ex}

        \begin{tikzpicture}
            \node[anchor=south west, inner sep=0] (image) at (0,0) { \includegraphics[width=\linewidth]{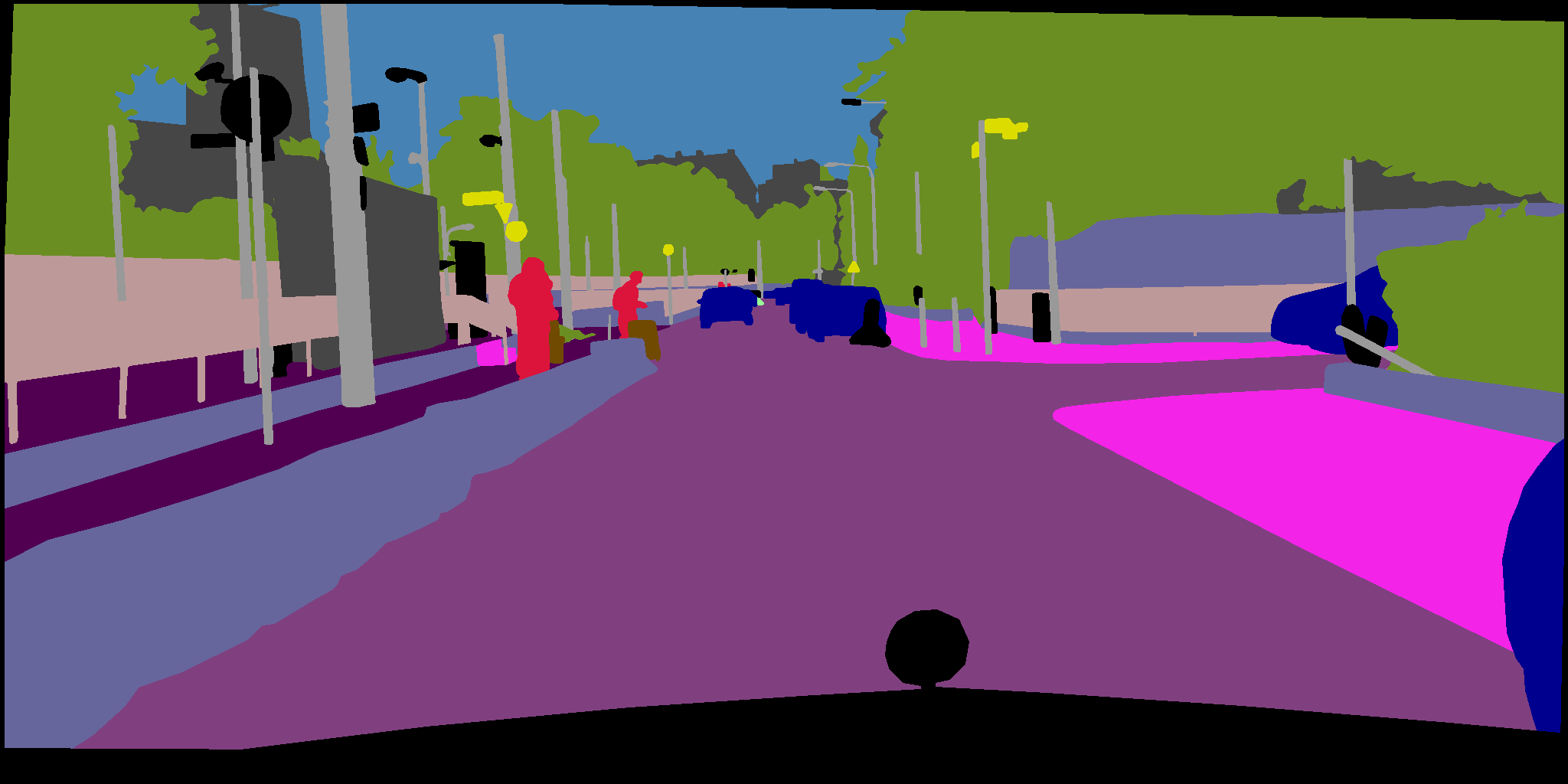} };
            \begin{scope}[x={(image.south east)}, y={(image.north west)}]
                \draw[white, dashed] (0.05, 0.25) rectangle (0.3, 0.6);  
            \end{scope}
            \end{tikzpicture}
        \vspace{0.5ex}
    \end{minipage}

    \vspace{-10px}
    \caption{\textbf{Qualitative Results: Synthia$\to$Cityscapes.} Existing DASS approaches face difficulty discerning visually similar classes (e.g. \textit{road} and \textit{sidewalk}). Our proposed method, LangDA, discerns classes with similar pixels while vision-only method struggles to do so.}
    \label{fig:qual_res_cityscapes}
\end{figure*}
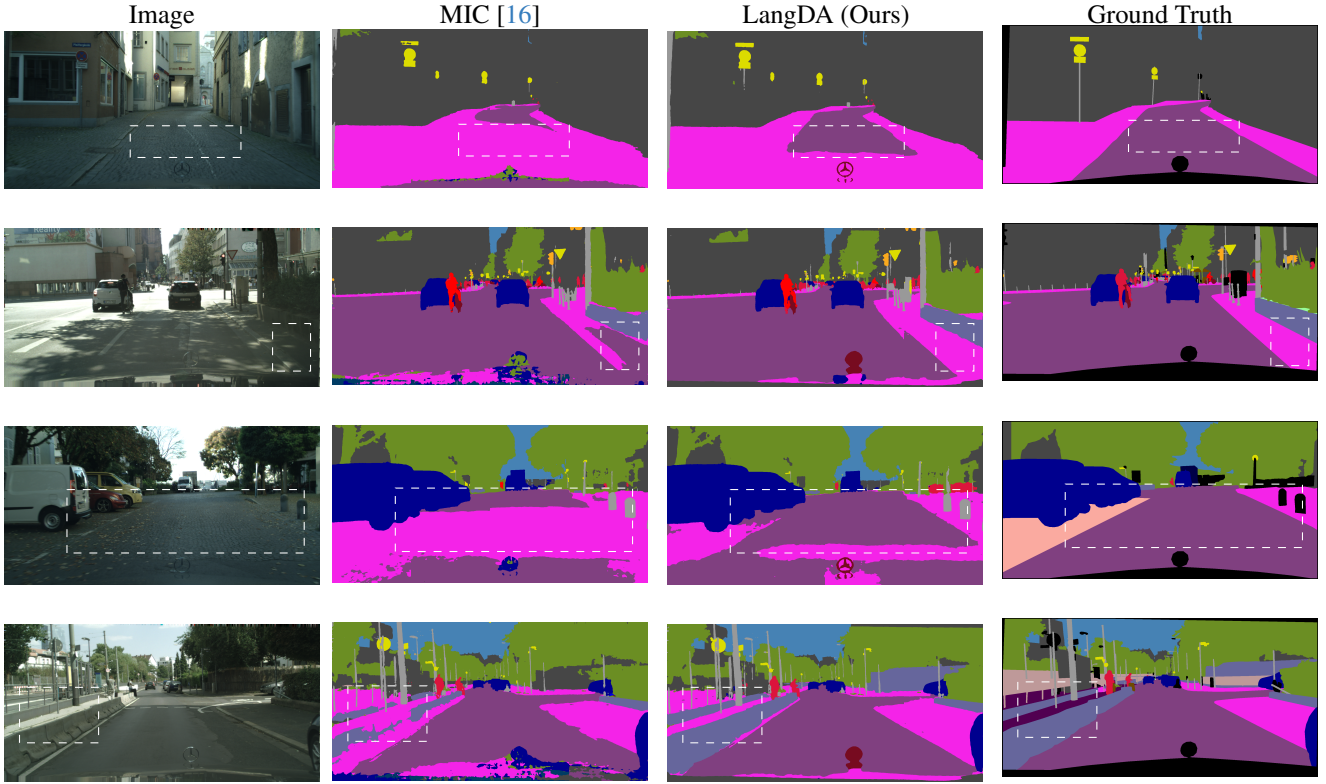

We provide a detailed class-wise comparison of existing methods with LangDA in Table \ref{tab:synthia_per_class_seg_iou}.  From \Cref{tab:synthia_per_class_seg_iou}, LangDA consistently achieves new SOTA on almost all classes. Specifically, LangDA significantly outperforms existing works on the \textit{sidewalk} and \textit{road} classes by 14\% and 8\% respectively. As seen in \Cref{fig:qual_res_cityscapes}, vision-only methods that try to build context-awareness \cite{hoyer2023mic} still struggle to differentiate classes with similar visual appearances (such as \textit{road} and \textit{sidewalk}). With the proposed image-level alignment module, the image features are aligned with textual features of the entire scene, facilitating contextual understanding of the difference between classes via language. These results reinforce the benefit of proposed image-level alignment module and building context through language. 

Additionally, LangDA outperforms existing methods in classes that are more reliant on context from 0.5\% to 2.87\%. For instance, the class \textit{rider} usually appears alongside bicycles, \textit{fence} is almost always in front of buildings, and \textit{pole} is often beside the road \cite{hoyer2023mic}. This result provides evidence that visual-only contextual alignment is insufficient in bridging domain gaps. Our proposed LangDA method captures additional contextual cues embedded in the latent prior of VLMs.  Moreover, our method occasionally benefits classes such as \textit{building} and \textit{vegetation} where context might not play a big role \cite{hoyer2023mic}. This showcases the significance of language world priors in our representations in improving such classes. See Appendix \ref{sec:tsne} for additional t-SNE visualization.

\subsubsection{Per-class IoU comparison on other two settings}
To evaluate LangDA's generalizability, we test it on two challenging scenarios: normal-to-adverse weather (Cityscapes $\to$ ACDC) in Table \ref{tab:cs2acdc_per_class_seg_iou} and day-to-night (Cityscapes $\to$ DarkZurich) in Table \ref{tab:cs2dz_per_class_seg_iou}. LangDA outperforms existing methods by 3.9\% and 1.4\% mIoU, respectively, setting a new state-of-the-art. However, similar to CoPT and MIC, we struggles with the \textit{sky} and \textit{vegetation} classes, where the model sometimes misclassifies the sky as trees for each other. This is likely due to improved segmentation from the source-trained domain, where trees frequently obscure the sky in the Cityscapes dataset. Additionally, ground truth labels for sky and trees are likely unreliable in DarkZurich due to the inherent difficulty for humans to distinguish trees from sky in low-light conditions. Despite this, LangDA consistently attain top one or two performances across most classes in both datasets. Notably, on CS $\to$ DZ, it achieves substantial improvements of 8.16\% and 8.74\% mIoU on the \textit{bus} and \textit{motorbike} classes, respectively, outperforming CoPT and MIC. This highlights the importance of context relations, especially for smaller classes that occupy less of the image, where understanding contextual positioning is crucial.
 
For CS $\to$ ACDC, LangDA outperforms existing SOTA by 3.9\% in mIoU as seen in Table \ref{tab:acdc-condition}. LangDA substantially outperforms MIC and CoPT on the fog and snow conditions (8.5\% and 7.5\% respectively) while maintaining a similar performance on the other two conditions. This is mainly attributed to the fact that fog and snow introduce uniform domain shifts primarily in terms of texture and visibility, rather than drastic and uneven changes in lighting or color. However, a setting like night introduces uneven sources of lighting (streetlights) when compared to a daytime urban dataset like Cityscapes. Since CoPT uses domain knowledge of the target data, they have slightly better performance in such settings. This shows that our domain-agnostic objective is more useful for uniform domain shifts.   

\begin{figure*}[t]
    \centering
    \begin{subfigure}{0.37\textwidth}
        \begin{tikzpicture}
            \node[anchor=south west, inner sep=0] (image) at (0,0) {        \includegraphics[width=\textwidth]{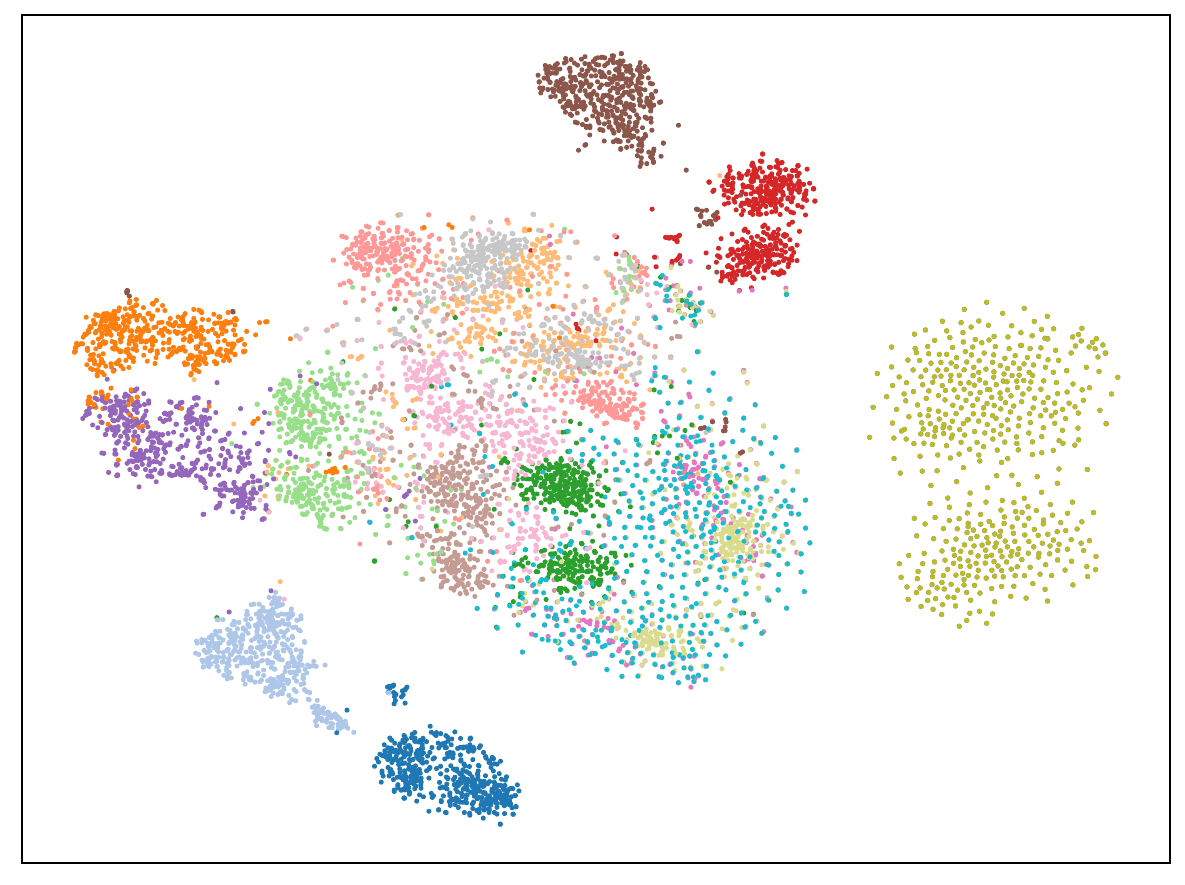}};
            \begin{scope}[x={(image.south east)}, y={(image.north west)}]
                \draw[black, dashed] (0.27, 0.5) rectangle (0.57, 0.77);
            \end{scope}
        \end{tikzpicture}
        \caption{\textbf{Left}: DAFormer \cite{hoyer2022daformer}, adaptation using only visual images.}
        \label{fig:tsne_daformer}
    \end{subfigure}
    \begin{subfigure}{0.37\textwidth}
        \begin{tikzpicture}
            \node[anchor=south west, inner sep=0] (image) at (0,0) {        \includegraphics[width=\textwidth]{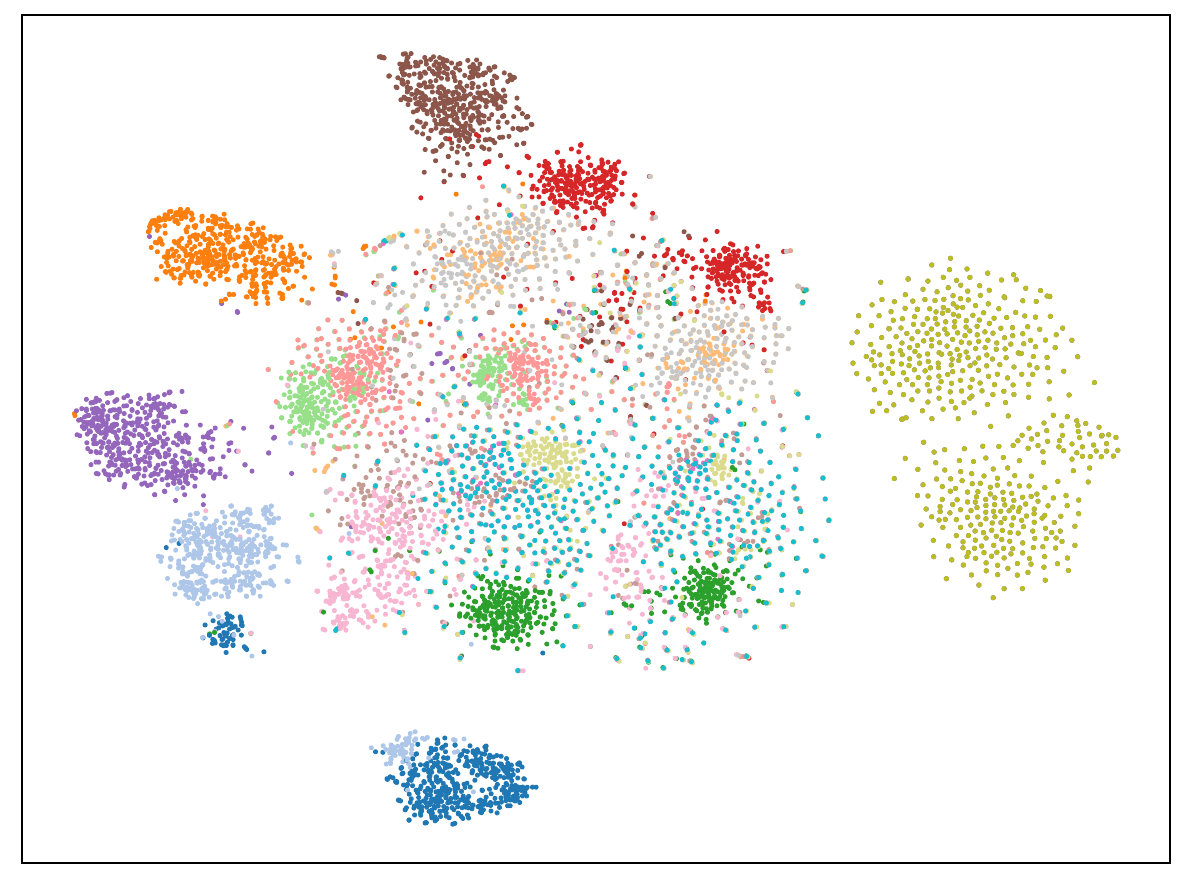}};
            \begin{scope}[x={(image.south east)}, y={(image.north west)}]
                \draw[black, dashed] (0.27, 0.5) rectangle (0.66, 0.8);
            \end{scope}
        \end{tikzpicture}
        \caption{\textbf{Right}: LangDA + DAFormer (Ours), adaptation using both visual images and contextual language descriptions.}
        \label{fig:tsne_LangDA}
    \end{subfigure}
    \begin{subfigure}{0.13\textwidth}
        \raisebox{0.2\textwidth}{
            \includegraphics[width=\textwidth]{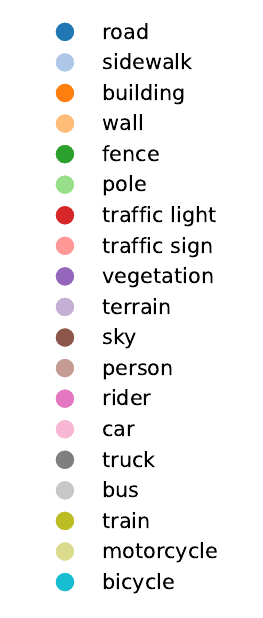}
        }
    \end{subfigure}
    \caption{\textbf{t-SNE of DAFormer and LangDA (Ours)} After aligning language and visual features, we observe more well-defined boundaries and improved class clustering. 
    }
    \label{fig:tsne_synthia}
\end{figure*}

\begin{table}[t!]
    \centering
    \caption{Effect of prompting and aligning techniques on Synthetic-to-Real adaptation benchmark: Synthia $\to$ Cityscapes.}
    \vspace{-10pt}
    \label{tab:ablation-prompt-align-2}
    \resizebox{\columnwidth}{!}{%
    \begin{tabular}{p{0.5cm}p{1cm}p{1cm}p{1cm}p{1cm}p{2.5cm}} \\
    \toprule
    \multicolumn{1}{c}{\textbf{}} & \multicolumn{2}{c}{Caption Generation} & \multicolumn{2}{c}{Alignment Technique} & \multicolumn{1}{c}{\textbf{\% mIoU$\uparrow$}} \\
    \cmidrule(r){2-3} \cmidrule(r){4-5}
    & \multicolumn{1}{c}{Context-aware Captions} & \multicolumn{1}{c}{Class-level Prompts} & \multicolumn{1}{c}{Image-level Alignment} & \multicolumn{1}{c}{Pixel-level Alignment} &  \\
    \midrule
    \multicolumn{1}{c}{1} & \multicolumn{1}{c}{\cmark} & \multicolumn{1}{c}{\xmark} & \multicolumn{1}{c}{\cmark} & \multicolumn{1}{c}{\xmark} & \multicolumn{1}{c}{\cellcolor[HTML]{B7E1CD}{\textbf{\synMiou}}} \\
    \multicolumn{1}{c}{2} & \multicolumn{1}{c}{\xmark} & \multicolumn{1}{c}{\cmark} & \multicolumn{1}{c}{\cmark} & \multicolumn{1}{c}{\xmark} & \multicolumn{1}{c}{68.9} \\
    \multicolumn{1}{c}{3} & \multicolumn{1}{c}{\xmark} & \multicolumn{1}{c}{\cmark} & \multicolumn{1}{c}{\xmark} & \multicolumn{1}{c}{\cmark} & \multicolumn{1}{c}{68.7} \\
    \multicolumn{1}{c}{4} & \multicolumn{1}{c}{\xmark} & \multicolumn{1}{c}{\xmark} & \multicolumn{1}{c}{\xmark} & \multicolumn{1}{c}{\xmark} & \multicolumn{1}{c}{67.3} \\
    \bottomrule
    \end{tabular}
}
\vspace{-15pt}
\end{table}


\begin{table}[!b]
\centering
\begin{minipage}{0.7\columnwidth}
    \centering
    \caption{Ablation: contextual caption applied on source, target, and source $+$ target}
    \vspace{-8px}
    \resizebox{\linewidth}{!}{%
    \begin{tabular}{ccc}
    \toprule 
    \bf Method & \bf Image Captions & \bf \% mIoU$\uparrow$ \\
    \midrule
    \rowcolor[HTML]{EFEFEF} MIC \cite{hoyer2023mic} & {\xmark} & 67.3\\
    \rowcolor[HTML]{EFEFEF} CoPT \cite{mata2024CoPT} & Source only & 67.4\\
    LangDA (Ours) & Source only & \cellcolor[HTML]{B7E1CD}{\textbf{\synMiou}}\\
    LangDA (Ours) & Target only & \cellcolor[HTML]{B7E1CD}{69.1}\\
    LangDA (Ours) & Source $+$ Target & \cellcolor[HTML]{B7E1CD}{68.0}\\
    \bottomrule
    \end{tabular}%
    }
    \label{tab:ablations-synthia-2}
\end{minipage}
\vspace{-10pt}
\end{table}

\subsubsection{t-SNE Visualization} \label{sec:tsne}
\cref{fig:tsne_synthia} shows the t-SNE visualizations of feature distributions of vision-only method DAFormer and LangDA built on top of it. After integrating language-driven feature alignment, our method LangDA shows improved per-class clustering. For instance, in DAFormer's \cite{hoyer2022daformer} t-SNE, the feature representations of walls (light orange) and traffic signs (rose pink) overlap in the image domain, likely due to traffic signs often visually appear in front of walls from driver's first-person view. On the other hand, walls and traffic signs are semantically distinguishable and do not appear together in language contextual descriptions, contributing to improved mIoU segmentation for LangDA in Table 2 of the main paper. These findings demonstrate the importance of inducing contextual relationships in language and highlight the advantages of LangDA for segmenting classes in UDA settings.

\subsection{Ablation Studies}

\textbf{Effect of Prompting and Alignment Strategies.} 
To evaluate the impact of the proposed modules, we replace context-aware caption generator with generic prompts and swap image-level alignment with pixel-level alignment in Tab. \ref{tab:ablation-prompt-align-2}. To assess the benefit of context-aware caption generator (row 2), we replace our contextual caption with text embedding of generic class-level prompts aligned to the entire image. This results in a 1.1\% drop in mIoU, demonstrating the effectiveness of the caption generator. To evaluate image-level alignment (row 3), we align generic class prompts (e.g. "A photo of a \{class\}") to the pixel features, leading to a 1.3\% mIoU reduction, showing the effectiveness of image-level alignment. While the gain from image-level alignment alone may seem modest (row 2 and 3), combining image-level alignment with context-aware caption generator (row 1 and 4) produces a substantial boost, demonstrating the exceptional synergy of our proposed modules.

\begin{table}[!b]
\centering
\begin{minipage}{0.67\columnwidth}
    \centering
    \vspace{-10px}
    \caption{Ablation: CLIP-based text encoders.}
    \vspace{-8px}
    \resizebox{\linewidth}{!}{%
    \begin{tabular}{ccc}
    \toprule 
    \multicolumn{1}{c}{\bf Method} & \multicolumn{1}{c}{\bf Text Encoder} & \multicolumn{1}{c}{\bf \% mIoU$\uparrow$} \\
    \midrule
    \rowcolor[HTML]{EFEFEF} MIC \cite{hoyer2023mic} & {\xmark} & 67.3\\
    \rowcolor[HTML]{EFEFEF} CoPT \cite{mata2024CoPT} & CLIP & 67.4\\
    LangDA (Ours) & OpenCLIP & \cellcolor[HTML]{B7E1CD}{68.9}\\
    LangDA (Ours) & LongCLIP & \cellcolor[HTML]{B7E1CD}{69.0}\\
    LangDA (Ours) & CLIP & \cellcolor[HTML]{B7E1CD}{\textbf{\synMiou}}\\
    \bottomrule
    \end{tabular}%
    }
    \label{tab:ablations-clip-synthia}
    \vspace{-10pt}
\end{minipage}
\hfill
\begin{minipage}{0.29\columnwidth}
    \centering
    \vspace{-10px}
    \caption{$\lambda_p$}
    \vspace{-8px}
    \resizebox{\linewidth}{!}{%
    \begin{tabular}{cc}
    \toprule
    \textbf{$\lambda_p$} & \textbf{\% mIoU $\uparrow$}\\
    \midrule
    2 & 67.9\\
    1 & 69.6\\
    0.1 & \textbf{\synMiou} \\
    0.01 & 69.7\\
    \rowcolor[HTML]{EFEFEF}{\xmark}  & 67.3 \\
    \bottomrule
    \end{tabular}%
    }
    \label{tab:weight_prompt_loss_synthia}
    \vspace{-10pt}
\end{minipage}
\end{table}

\noindent\textbf{Comparison of Source and Target Captions.}
In our main experiments, we use image-level captions on source data for text-based supervision. Table \ref{tab:ablations-synthia-2} examines whether applying our objective to target image captions can provide further improvement. For unsupervised target captions, we omit ground-truth masks, as these are considered unavailable. Instead, we prompt the LLM to summarize the scene description without further class name refinement, while the VLM receives text query not containing class names. The table shows that applying LangDA on either source or target captions achieves state-of-the-art (SOTA) performance on Synthia $\to$ CS, supporting our hypothesis that \textit{explicit language-induced context is advantageous}. The slight performance drop for target-only supervision may be attributed to the hallucinations of VLMs and LLMs without class name refinement. Furthermore, similar to CoPT \cite{mata2024CoPT}, when applying our objective to both source and target data, we backpropagated once after applying it only on source data to avoid memory issues, which alters the optimization process and may lead to lower performance.

\noindent\textbf{Hyperparameter Sensitivity}
\label{sec: hyper_sensi}
LangDA exhibits low sensitivity to $\lambda_p$, weight of the language consistency loss, eliminating the need for extensive hyperparameter tuning. As shown in \Cref{tab:weight_prompt_loss_synthia}, performance remains strong as long as $\lambda_p$ remains within a reasonable range and is weighted less than the combined unsupervised and supervised losses. LangDA's robustness to hyperparameter changes makes it simple and practical to deploy.

\noindent\textbf{Ablations on CLIP Text Encoders} 
\label{sec:abla_clip}
To study LangDA's dependence on text encoder, we generate text embeddings by passing our refined context-aware captions through different text encoders \cite{radford2021clip,openclip} and retrain LangDA using each of these text embeddings individually for Synthia $\to$ Cityscapes (CS). The results are reported in \Cref{tab:ablations-clip-synthia}. From the table, it is evident that LangDA outperforms MIC \cite{hoyer2023mic} (67.3\%) and CoPT \cite{mata2024CoPT} (67.4\%) irrespective of the chosen text encoder, highlighting the significance of the proposed context-aware caption generator. 





\section{Conclusion}
\label{sec:conclusion}

In this paper, we introduce LangDA, the first work to explicitly induce spatial relationships through language to learn rich context relations and generalized representations for DASS. LangDA achieves this by developing a) a context-aware caption generator that expresses spatial context in images using language, and b) an image-level alignment module that promotes context awareness via alignment between the scene caption and the entire image. Unlike prior methods relying on pixel-level alignment, LangDA’s image-level consistency aligns with the entire image, promoting a holistic understanding of class relationships. Notably, LangDA is also the first language-guided DASS method that requires no supervisory text for describing the target domain (e.g. "a \{daytime\} photo" to "a \{snowy\} photo"). As shown in extensive experiments and ablations, LangDA establishes a new SOTA on three key DASS benchmarks, achieving significant performance improvements over existing methods and consistently improving performance when integrated with other UDA methods. In doing so, LangDA establishes that extracting key context relationships from language is a promising avenue for DASS. Moreover, it also motivates future work in automatically tailoring captions to draw critical information for a particular task.



\clearpage
{
    \small
    \bibliographystyle{ieeenat_fullname}
    \bibliography{main}
}

\clearpage
\setcounter{page}{1}
\maketitlesupplementary

\begin{figure}[t]
    \includegraphics[width=\columnwidth]{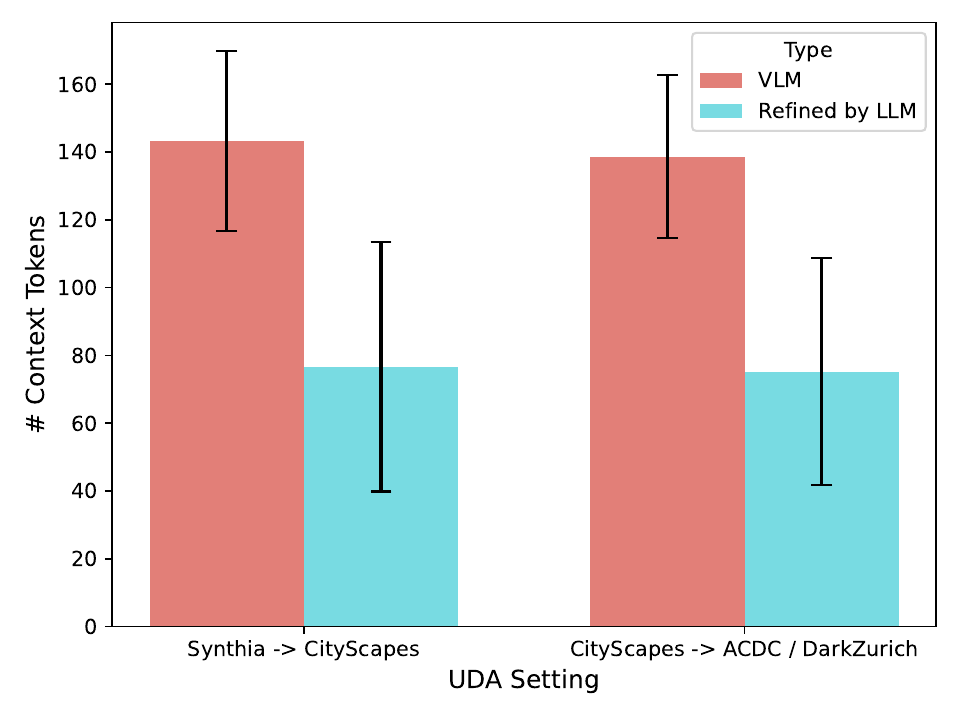}%
    \vspace{-5px}
    \caption{\textbf{Token Length After Refinement on Synthia and Cityscapes.} Token length is shorter after refinement (an average of around 77 tokens, the maximum accepted token length of CLIP).
    }
    \label{fig:token_len_refine}
    \vspace{-15px}
\end{figure}

\begin{figure*}[b]
    \centering
    \begin{subfigure}{0.45\textwidth}
        \includegraphics[width=\columnwidth]{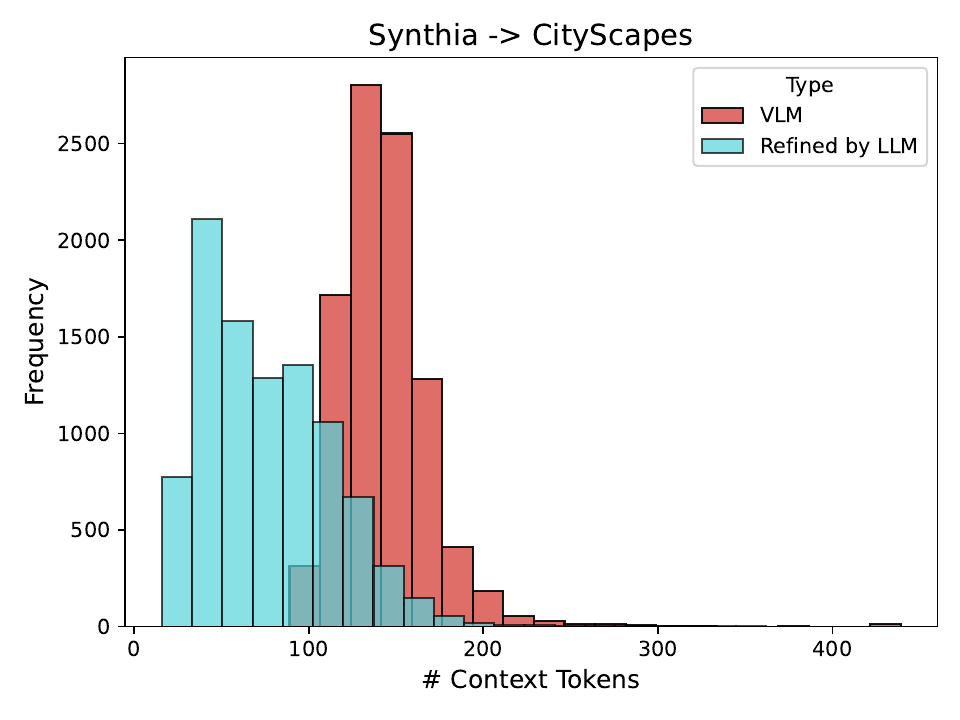}
        \label{fig:token_len_hist_syn}
    \end{subfigure}
    \begin{subfigure}{0.45\textwidth}
        \includegraphics[width=\columnwidth]{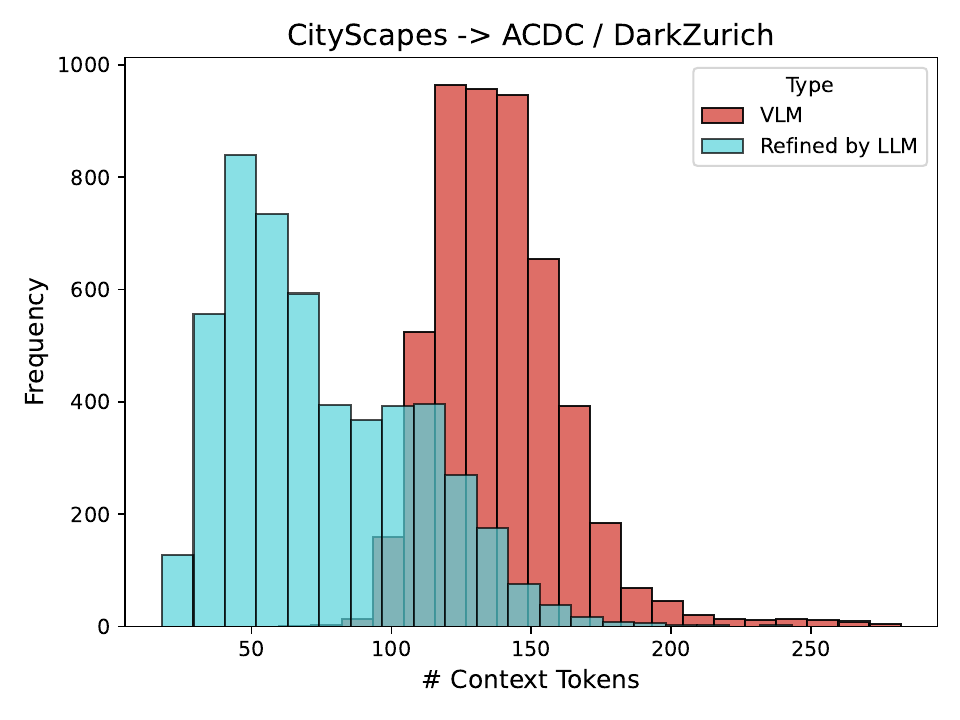}
        \label{fig:token_len_hist_cs}
    \end{subfigure}
    \vspace{-10px}
    \caption{
        \textbf{Left}: Token Length Distribution on Synthia. Token length for VLM is much longer than token length for LLM after refinement. Most token length centered around 150 before refinement, whereas token length centered around 70 after refinement. \textbf{Right}: Token Length Distribution on Cityscapes. Note the distribution of token length is much longer and less uniform compared to Synthia. This is because Cityscapes is real-world data, which contains varied scenes and requires more detailed description.
    }
    \label{fig:context_syncs}
\end{figure*}



\section{Quality of Image-level Descriptors} \label{sec:qual_img_desc}

\cref{fig:token_len_refine} shows the tokens before refinement have average context lengths of around 140 tokens on both Synthia and Cityscapes, which is too long for CLIP \cite{radford2021clip} context length that only accepts a maximum of 77 tokens. After refinement, we can see the average context length drops down to around 77 tokens for both datasets, showcasing the necessity of our LLM-based caption refinement process. This is further elucidated in \cref{fig:context_syncs}, where the distributions of context lengths of raw and refined captions are visualized. From this figure, we can clearly see that the distributions of context lengths of the refined captions for both datasets are centered around 70. These results, combined with the example refined prompt in Fig. 4 from the main paper, demonstrate that the LLM refinement module reduced the number of tokens significantly while preserving the context relationships.

\section{Additional Related Works: Unsupervised Domain Adaptation (UDA)}
\label{sec:additional_rel_works}
In UDA, a model trained on a labeled source domain is adapted to an unlabeled target domain. Most UDA approaches rely on discrepancy minimization \cite{long2017maxmeandiscre, sun2016corr_al,sun2016corr_return, vu2019advent,grandvalet2004semi_entrop_min}, adversarial training \cite{vu2019advent,goodfellow2014gan}, or self-training \cite{chen2022deliberated,hoyer2022daformer,kim2023bidirectional,lee2013pseudo,tarvainen2017mean_tea_consis_regu}.
Discrepancy minimization involves reducing the domain gap using statistical distance functions like maximum mean discrepancy \cite{long2017maxmeandiscre}, correlation alignment \cite{sun2016corr_al,sun2016corr_return}, or entropy minimization \cite{vu2019advent,grandvalet2004semi_entrop_min}. Adversarial training uses a learned domain discriminator within a GAN framework \cite{goodfellow2014gan} to promote domain-invariant inputs, features, or outputs \cite{vu2019advent}. Self-training generates pseudo-labels \cite{lee2013pseudo} for the target domain based on confidence thresholds \cite{hoyer2022daformer}, with consistency regularization \cite{tarvainen2017mean_tea_consis_regu, tranheden2021dacs} often applied to ensure robustness across different data augmentations \cite{hoyer2023mic,araslanov2021uda_data_aug} and domain-mixup \cite{olsson2021classmix,yun2019cutmix,kim2023bidirectional, tranheden2021dacs}. Other works also apply consistency regularization using prototypes \cite{zhang2021prototypical} and auxiliary task correlation \cite{wang2021domain}.

\end{document}